\documentclass[10pt,journal,compsoc]{IEEEtran}
\usepackage{url}
\usepackage{graphicx}
\usepackage{amsmath}
\usepackage{amssymb}
\usepackage{algorithmic}
\usepackage[tight,footnotesize]{subfigure}
\usepackage{color}
\usepackage[colorlinks=true,citecolor=red]{hyperref}
\usepackage{multirow}

\def\win{\color{red}\bf}

\newtheorem{definition}{Definition}

\newtheorem{proposition}{Proposition}
\newtheorem{lemma}{Lemma}
\newtheorem{proof}{Proof}
\newtheorem{property}{Property}

\def\imheight{2.5cm}

\ifCLASSOPTIONcompsoc
  \usepackage[nocompress]{cite}
\else
  \usepackage{cite}
\fi
\ifCLASSINFOpdf
\else
\fi

\begin{document}
%
\title{Efficient Discriminative Nonorthogonal Binary Subspace with its Application to Visual Tracking}
%
%
%
%

\author{Ang~Li,
        Feng~Tang,
        Yanwen~Guo,
        and~Hai~Tao
\IEEEcompsocitemizethanks{\IEEEcompsocthanksitem A. Li is with the Department
of Computer Science and the Institute for Advanced Computer Studies, University of Maryland, College Park,
MD, 20742. E-mail: angli@umiacs.umd.edu.

\IEEEcompsocthanksitem F. Tang is with the Hewlett-Packard Laboratories, 1501 Page Mill Rd, Palo Alto, CA 94304, USA. Email: sharp.tang@gmail.com.
\IEEEcompsocthanksitem Y. Guo is with the National Key Laboratory for Novel Software Technology, Nanjing University, Nanjing, China 210023, and the
              Coordinated Science Lab, University of Illinois, Urbana Champaign, IL 61801. Email:~ywguo@nju.edu.cn.
              \IEEEcompsocthanksitem H. Tao is with the Department of Computer Engineering, University of California, Santa Cruz, CA 95064. Email:~taohai2006@gmail.com.}
              }

%
%

\markboth{Extended version of Nonorthogonal Binary Subspace Tracking ECCV 2010}%
{Li \MakeLowercase{\textit{et al.}}: Efficient Discriminative Nonorthogonal Binary Subspace with its Application to Visual Tracking}
%



\IEEEtitleabstractindextext{%
\begin{abstract}
One of the crucial problems in visual tracking is how the object is represented.
Conventional appearance-based trackers are using increasingly more
complex features in order to be robust. However, complex
representations typically not only require more computation for
feature extraction, but also make the state inference complicated.
We show that with a careful feature selection scheme,
extremely simple yet discriminative features can be used for robust
object tracking. The central component of the proposed method is a
succinct and discriminative representation of the object using
discriminative non-orthogonal binary subspace (DNBS) which is spanned by Haar-like
features. The DNBS representation inherits the merits of the
original NBS in that it efficiently describes the
object. It also incorporates the discriminative information to
distinguish foreground from background. However, the problem of finding the DNBS bases from an over-complete dictionary is NP-hard. We propose a greedy algorithm called discriminative optimized orthogonal
matching pursuit (D-OOMP) to solve this problem. An iterative formulation named iterative D-OOMP is further developed to drastically reduce the redundant computation between iterations and a hierarchical selection strategy is integrated for reducing the search space of features.  The proposed DNBS representation is applied to object tracking through SSD-based template
matching. We validate the
effectiveness of our method through extensive experiments on
challenging videos  with comparisons against several state-of-the-art trackers and demonstrate its capability to track objects
in clutter and moving background.
\end{abstract}

}

\maketitle

\IEEEdisplaynontitleabstractindextext

%
\IEEEpeerreviewmaketitle

\IEEEraisesectionheading{\section{Introduction}\label{intro}}
\IEEEPARstart{V}{isual} object tracking in video sequences is an active research
topic in computer vision, due to its wide applications in video
surveillance, intelligent user interface, content-based video
retrieval and object-based video compression. Over the past two
decades, a great variety of tracking methods have been brought
forward. Some of them include template/appearance based methods
\cite{Hager-04,Han-05,Matthews-pami-04,Black-eccv-96,Cootes-pami-01},
layer based methods \cite{Jepson-pami-03, Tao-pami-02}, image
statistics based methods
\cite{Comaniciu-pami-03,Fan-05,Birchfield-05}, feature based methods
\cite{Shi-94,Tang-05}, contour based methods \cite{Chen-cvpr-01},
and discriminative feature based methods
\cite{Collins-pami-05,Avidan-pami-05}. One of the most popular
categories of methods is appearance based approaches which
represent the object to be tracked using an appearance model and match the model
 to each new frame to determine the object state. In order
to handle appearance variations, an appearance update scheme is
usually employed to adapt the object representation over time.
Appearance based trackers have shown to be very successful in many
scenarios. However they may not be robust to background clutter
where the object is very similar to background. In order to
solve this problem, more and more complicated object
representations which take into account colors, gradients and textures are
used. However, extraction of the complicated features usually incurs
more computation which slows down the tracker. Moreover, complex
representation will make the inference much more complicated. One
natural question to ask is how complicated features are really needed
to track an object. In this paper, we show that with a careful
feature selection scheme, extremely simple object representations
can be used to robustly track objects.

Essentially, object tracking boils down to the image representation
problem -- what type of feature should be used to represent an
object? An effective and efficient image representation not only makes
the feature extraction process fast but also reduces computational load
for object state inference. Traditional object representations such as raw pixels and color histograms are generative in natural, which
are usually designed to describe the appearance of object being
tracked while completely ignoring the background. Trackers using
this representation may fail when the object appearance is very
similar to the background. It is worth noting that some appearance
based trackers model both foreground and background, for example in
the layer tracker \cite{Tao-pami-02} the per-pixel layer ownership
is inferred by competing the foreground and background likelihoods using a mixture of Gaussians. However the Gaussian model assumption degrades the representation power of the model.  Subspaces are popular in modeling the object appearance. IVT \cite{ivt} incrementally learns principal components of the object to adapt its appearance changes during tracking. Compressive Tracking \cite{ZhangCT} was proposed to project the object features into a subspace spanned by sparse binary basis. Most of these methods consider only the object appearance while not aware of the background context information.

Recently, discriminative approaches have opened a promising
new direction in the tracking literature by posing tracking as a
classification problem. Instead of trying to build an appearance
model to describe the object, discriminative trackers seek a
decision boundary that can best separate the object and background (such as \cite{svt, henriques2012circulant, Struck, TLD}).
The support vector tracker \cite{svt} (denoted as SVT afterwards)
uses an offline-learned support vector machine as the classifier and
embeds it into an optical flow based tracker. Recently, Struck \cite{Struck} employed the structural support vector machines to learn the object classifier and achieved the state-of-the-art performance.  TLD \cite{TLD} uses ferns as a rough classifier which generates object candidates and verifies these object patches using a nearest neighbor classifier. Collins et
al.~\cite{Collins-pami-05} are perhaps the first to treat tracking
as a binary classification problem.  A feature
selection scheme based on variance ratio is used to select the most
discriminative features for tracking in the next frame. Avidan's ensemble tracker
~\cite{Avidan-pami-05} combines an ensemble of online learned weak
classifiers using AdaBoost to classify pixels in the new frame. In discriminative spatial attention tracking \cite{fan-eccv10}, attention regions (AR) which are locally different from their neighborhoods are selected as discriminative tracking features. In \cite{nguyen-ijcv}, Gabor features are used to represent an object and the background. A differential version of Linear Discriminant Analysis classifier is built and maintained for tracking. In these trackers, the tracking result in the current frame is usually used to select training samples to update the classifier. This bootstrap process is sensitive to tracking errors -- slight inaccuracies can lead to incorrectly labeled training examples, hence degrading the classifier and finally causing further drift. To solve this problem, researchers have proposed to use more robust learning algorithms such as semi-supervised learning and multiple instance learning to learn from uncertain data. In co-tracking~\cite{Tang-iccv-07}, two semi-supervised support vector machines are built using color and gradient features to jointly track the object using co-training. In the online multiple instance tracking \cite{babenko-cvpr-09}, the classifier is learned using multiple instance learning which only requires bag-level labels so that makes the learner more robust to localization errors. Leistner et al. \cite{Christian-eccv-10} use online random forest for multiple instance which achieves faster and more robust tracker. In \cite{Zeisl-cvpr-10}, the authors combine multiple instance learning and semi-supervised learning in a boosting framework to minimize the propagation of tracking errors. The algorithm proposed in \cite{Saffari-cvpr-10} models the confusing background as virtual classes and solves the tracking problem in a multi-class boosting framework.

%

Previous discriminative trackers generally have two major problems.
First, the tracker only relies on the classifier which can well
separate the foreground and background and does not have any
information about what the object looks like. This makes it hard to recover
once the tracker makes a mistake. Second, discriminative trackers
generally have a fixed set of features for all objects to be
tracked and this representation is not updated any more. However,
adaptive objective representation is more desirable in most cases
because it can capture the appearance variations of the particular object
being tracked.

In this paper, we propose an extremely simple object representation
using Haar-like features for efficient object tracking. The representation is
generative in nature in that it finds the features that can best
reconstruct the foreground object. It is also discriminative because
only those features that make the foreground representation
different from background are selected. Our representation is based
on the nonorthogonal binary subspace (NBS) method in \cite{nbs:pami}.
The original NBS tries to select from an over-complete set of Haar-like features that can best represent the image. We propose a novel discriminative representation called discriminative nonorthogonal binary subspace(D-NBS) that
extends the NBS method to incorporate discriminative information. The new representation inherits the merits of
original NBS in that it can be used to efficiently describe the
object. It also incorporates the discriminative information to
distinguish foreground from background. The problem of finding a D-NBS subspace for a given template is NP-hard and even achieving an approximate solution is time consuming. We also propose a hierarchical search method that can efficiently find the subspace representation for a given image or a set of images. To make the tracker more robust, an update scheme is devised in
order to accommodate object appearance variations and background change. We validate the
effectiveness of our approach through extensive experiments on
challenging videos and demonstrate its capability to track objects
in clutter and moving background.

It is worth noting that there are also methods in machine learning that combines generative and discriminative models, for example \cite{Jaakkola-nips-98, lin-nips-04, lasserre-cvpr-06, Raina-nips-03, eigenboost-cvpr-07, tu-cvpr-07, Saffari-cvpr-10}. Grabner et al. proposed to use boosting to select Haar-like features and these features are used to approximate a generative model \cite{eigenboost-cvpr-07}. Tu et al. proposed an approach to progressively learn a target generative distribution by using negative samples as auxiliary variables to facilitate learning via discriminative models \cite{tu-cvpr-07}. This idea has been widely applied in later computer vision literatures. In \cite{yu-eccv-08}, a generative subspace appearance model and a discriminative online support vector machine are used in the co-training framework for tracking objects. However, in this work, two different representations are used for generative and discriminative model. This would incur extra computation for feature extraction. In this work, we propose a principled method to extract a set of highly efficient features that are both generative and discriminative.
%

%
A preliminary version of this work appeared as a conference paper \cite{ang-eccv-2010}.
This paper extends the previous version in the following perspectives:
\begin{itemize}
\item We devise a novel iterative D-OOMP method for fast computation of the D-NBS representation. This iterative method exploits the redundancy between the iterations of feature selection with a recursive formulation, hence significantly reducing the computational load.
\item We propose a hierarchical D-OOMP algorithm that can speed up the search using a hierarchical dictionary obtained by feature clustering.  This process dramatically reduces the computation cost and makes our approach applicable to large templates.
\item We provide more detailed performance analysis and extensive experiments to show the superiority of the new method in this paper over the preliminary version of this work. We compare our tracker against 8 state-of-the-art trackers in 21 video sequences using comprehensive evaluation criteria.
\end{itemize}

The rest of this paper is organized as follows. In Section \ref{sec:nbs}, we
briefly review Haar-like features and the non-orthogonal binary
subspace approach. The Discriminative Nonorthogonal Binary Subspace (DNBS) formulation and its optimization algorithm (D-OOMP) are proposed in Section \ref{sec:dnbs}. Section \ref{sec:fast} introduces an equivalent DNBS formulation which speeds up the feature selection without loss of accuracy. Besides, a hierarchical strategy is further incorporated to boost the performance. In Section \ref{sec:track}, the application of
DNBS to tracking is described. Both qualitative and quantitative experimental results are given in Section \ref{sec:experiment}. Finally, we conclude the paper and discuss the future work in Section \ref{sec:conclusion}.

\section{Background: Nonorthogonal Binary Subspace} \label{sec:nbs}
Haar-like features and the variants have been widely employed in object detection and tracking
\cite{viola-face, Mita-iccv-05, Kalal2010, Grabner-cvpr-06, babenko-cvpr-09} due to its computational efficiency.
The original Haar-like features measure the intensity difference between black and white box regions in an image. This definition was modified in \cite{nbs:pami} as the sum of all the pixels in a white box region for the purpose of image reconstruction.

\begin{definition}[Haar-like function]
The Haar-like box function $\mathcal{H}$ for Nonorthogonal Binary Subspace is defined as,
\begin{equation}
\mathcal{H}_{u_0,v_0,w,h}(u,v)=\left\{
\begin{array}{ll}
1,&\quad u_0\le u\le u_0+w-1\\
&\quad v_0\le v\le v_0+h-1\\
0,&\quad\text{otherwise}~,
\end{array}\right.
\end{equation}
where $w$ and $h$ represent the width and height of the box in the
template. $(u_0,v_0)$ represents the top-left location of the Haar-like box. The advantage of such
box functions is that the inner product of the Haar-like base with
any same-sized image template can be computed with only 4 additions,
by pre-computing the integral image of the template.
\end{definition}

The original NBS \cite{nbs:pami} approach tries to find a subset of Haar-like
features from an overcomplete dictionary to span a subspace that can
be used to reconstruct the original image. It is worth noting that in \cite{nbs:pami} the Haar-like functions have two types,
i.e. one-box and symmetrical two-box functions. The symmetrical two-box functions are mainly designed for images with symmetric structure (e.g. frontal faces). We select only one-box functions to make it suitable for tracking arbitrary object that may not have symmetric structures.

Suppose that for any given image template $\mathbf{x}\in \mathbb{R}^{W \times H}$ of
size $W\times H$ and the selected binary box features are
$\{c_i,\phi_i\}$($1\le i\le K$). $c_i$ is the coefficient of box
function $\phi_i$. The NBS approximation is formulated as
$\mathbf{x}=\sum_{i=1}^Kc_i\mathbf{\phi}_i+\mathbf{\varepsilon}$,
where $\mathbf{\varepsilon}$ denotes the reconstruction error. We
define $\mathbf{\Phi}_K=[\phi_1,\phi_2,\ldots,\phi_K]$ as
the basis matrix, each column of which is a binary base vector. Note
that, this base set is non-orthogonal in general, therefore the
reconstruction vector of template $\mathbf{x}$ should be calculated by the Moore-Pense pseudo-inverse such that
\begin{equation}\label{nbs:recon}
R_{\mathbf{\Phi}_K}(\mathbf{x})=\mathbf{\Phi}_K(\mathbf{\Phi}_K^\mathsf{T}\mathbf{\Phi}_K)^{-1}\mathbf{\Phi}_K^\mathsf{T}\mathbf{x}~.
\end{equation}

\begin{definition}
For a given image template with width $W$ and height $H$, a nonorthogonal binary feature dictionary $\mathbf{D}_{W,H}$ is specified such that
\begin{align}
\mathbf{D}_{W,H}=\{&\mathcal{H}_{u_0,v_0,w,h}~|~u_0,v_0,w,h\ge 1\nonumber\\
&\wedge u_0+w-1\le W\wedge v_0+h-1\le H\}~.
\end{align}
\end{definition}
The dictionary is composed of all possible Haar-like box functions which vary by the location and size of the white box. In our formulation introduced later in this paper, we represents the dictionary using a matrix $\mathbf{\Psi}=[\psi_1,\psi_2,\ldots\psi_{N_\psi}]$ where each column vector is a vectorized Haar-like feature in dictionary $\mathbf{D}_{W,H}$. The total number of Haar-like box functions $N_\psi$ in dictionary $\mathbf{D}_{W,H}$ is $W(W+1)H(H+1)/4$, thus the
dictionary of base vectors is over-complete and highly redundant. The objective function for the optimal subspace selection with respect to a given image template is to minimize the reconstruction error using selected base vectors, which is formulated as
\begin{equation}\label{nbs:func}
\arg\min_{\mathbf{\Phi}_K}{\parallel\mathbf{x}-R_{\mathbf{\Phi}_K}(\mathbf{x})\parallel}~.
\end{equation}
In general, the problem of optimizing
Eq. \ref{nbs:func} is NP-hard. Greedy approximate solutions for example optimized orthogonal matching pursuit (OOMP) \cite{nbs:pami,oomp1} have been proposed to find a sub-optimal set of base vectors by iteratively selecting a base vector that minimizes the reconstruction error.

\section{Discriminative Nonorthogonal Binary Subspace}\label{sec:dnbs}
The NBS method has been successfully used in computer vision applications such as fast template matching \cite{nbs:pami}. However, we find it less robust for applications such as object tracking. This is because tracking is essentially a binary classification problem to distinguish between foreground and background. NBS only considers the
information embodied in the object image itself without any
information about the background. To solve this problem, we propose the
Discriminative Non-orthogonal Binary Subspace (D-NBS) image representation that extracts features using both
positive samples and negative samples, i.e. foreground objects and background. The discriminative NBS method
inherits the merits of the original NBS in that it can well describe
the object appearance, and at the same time, it captures the
discriminant information that can better separate the object from
background.

\subsection{Formulation}
The objective of Discriminative NBS is to construct an object
representation that can better distinguish between foreground object and background. The main idea behind Discriminative NBS is that we want to select
features so that the reconstruction error for foreground is small
while it is large for background. Different from the original NBS
formulation Eq. \ref{nbs:func} in which only the foreground
reconstruction error is considered, in Discriminative NBS formulation, the
objective function has both foreground and background reconstruction
terms.

\begin{figure}[!ht]
\centering
\includegraphics[width=.4\textwidth]{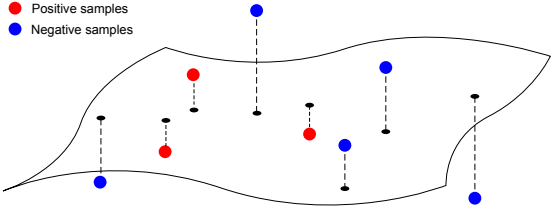}
\caption{An illustration of subspaces trained using multiple positive and negative samples: positive examples have smaller reconstruction errors while negative ones have larger reconstruction errors}
\end{figure}

Let $\mathbf{\Phi}_K$ be the Discriminative NBS basis vectors with
$K$ bases and $R_{\mathbf{\Phi}_K}(\mathbf{X})$ be the
reconstruction of $\mathbf{X}$ via $\mathbf{\Phi}_K$ using
Eq. \ref{nbs:recon}. Note that
$\mathbf{F}=\left[\mathbf{f}_1,\mathbf{f}_2,\ldots,\mathbf{f}_{N_f}\right]$
is a matrix of $N_f$ recent foreground samples and $\mathbf{B}=\left[\mathbf{b}_1,\mathbf{b}_2,\ldots,\mathbf{b}_{N_b}\right]$
is a matrix of $N_b$ sampled background vectors. The objective
function for $\mathbf{\Phi}_K$ is to optimize
\begin{equation}\label{dnbs:form}
\arg\min_{\mathbf{\Phi}_K}\frac{1}{N_f}\parallel\mathbf{F}-R_{\mathbf{\Phi}_K}(\mathbf{F})\parallel_F^2-\frac{\lambda}{N_b}\parallel\mathbf{B}-R_{\mathbf{\Phi}_K}(\mathbf{B})\parallel_F^2
\end{equation}
where $\parallel\cdot\parallel_F$ represents the Frobenius norm. The
first term in the equation is the reconstruction error for the foreground and the second term is the reconstruction error for the background. The objective is to find the set of base vectors to minimize the foreground reconstruct error while maximizing the background errors. This formulation can be interpreted as a hybrid form in which
the generative and discriminative items are balanced by $\lambda$. As the feature dictionary is highly redundant and over-complete, the original NBS is under constrained. The second discriminative term can also be viewed as a regularization term to constrain the solution.
An equivalent formulation of Eq. \ref{dnbs:form} is
\begin{equation}\label{dnbs:form2}
\arg\max_{\mathbf{\Phi}_K}\frac{1}{N_f}\sum_{i=1}^{N_f}\langle\mathbf{f}_i,R_{\mathbf{\Phi}_K}(\mathbf{f}_i)\rangle-\frac{\lambda}{N_b}\sum_{i=1}^{N_b}\langle\mathbf{b}_i,R_{\mathbf{\Phi}_K}(\mathbf{b}_i)\rangle~.
\end{equation}

Conventional discriminative tracking approaches only model the difference between foreground and background, so they hardly memorize information about what the object looks like. Once losing track, they have weaker ability to recover, compared to generative trackers. The proposed approach has a generative component of object appearance, i.e., the model constrains the tracked result to be similar to the object in appearance. Such an enhanced model reduces the chance of losing track. In addition, such a combined generative-discriminative approach also helps recover the object from tracking failure. 

\subsection{Solution: Discriminative OOMP}
It can be proved that solving Eq. \ref{dnbs:form} is NP hard, even
verification of a solution is difficult. To optimize it, we propose an extension of OOMP (Optimized Orthogonal
Matching Pursuit) \cite{nbs:pami} called discriminative OOMP.
Similar to OOMP, discriminative OOMP is a greedy algorithm which
computes adaptive signal representation by iteratively selecting
base vectors from a dictionary.

We assume that totally $K$ base vectors are to be chosen from the
dictionary $\mathbf{\Psi}=[\psi_1,\psi_2,\ldots,\psi_{N_\psi}]$ where
$N_\psi$ is the total number of base vectors in the dictionary.
Supposing $k-1$ bases
$\mathbf{\Phi}_{k-1}=[\phi_1,\phi_2,\ldots,\phi_{k-1}]$ have been
selected, the $k$-th base is chosen to best reduce the construction errors for foreground and least for the background. Note that the candidate feature $\phi_i$ may not be orthogonal to the subspace $\mathbf{\Phi}_{k-1}$, the real contribution of $\psi_i$ to increase Eq. \ref{dnbs:form2} has to be offset by the component that lies in $\mathbf{\Phi}_{k-1}$.  So the objective is to find the $\psi_i$ which maximizes the following function:
\begin{equation}\label{eq:doomp}
\frac{1}{N_f}\sum_{j=1}^{N_f}\frac{|\langle\gamma_i^{(k)},\varepsilon_{k-1}(\mathbf{f}_j)\rangle|^2}{\parallel\gamma_i^{(k)}\parallel^2}
-\frac{\lambda}{N_b}\sum_{j=1}^{N_b}\frac{|\langle\gamma_i^{(k)},\varepsilon_{k-1}(\mathbf{b}_j)\rangle|^2}{\parallel\gamma_i^{(k)}\parallel^2}
\end{equation}
where
$\gamma_i^{(k)}=\mathbf{\psi}_i-R_{\mathbf{\Phi}_{k-1}}(\mathbf{\psi}_i)$
is the component of base vector $\psi_i$ that is orthogonal to the
subspace spanned by $\mathbf{\Phi}_{k-1}$.
$\varepsilon_{k-1}(\mathbf{x}) =
\mathbf{x}-R_{\mathbf{\Phi}_{k-1}}(\mathbf{x})$ denotes the
reconstruction error of $\mathbf{x}$ using $\mathbf{\Phi}_{k-1}$.

In each iteration of the base selection, the algorithm needs to
search all the dictionary $\mathbf{\psi}_i$ to compute
$\gamma_i^{(k)}$. Since the number of bases in dictionary is
quadratic to the number of pixels in image, this process may be slow
for large templates. To solve this problem, we further analyze the components of the above equation for
for recursive formulation for fast computation.

\begin{property}[Inner product]\label{prop1} Let $\mathbf\Phi$ be a subspace in $\mathbb R^n$. For any point $\mathbf x,\mathbf y\in \mathbb R^n$, 
\begin{equation}
\langle\mathbf x - R_{\mathbf\Phi}(\mathbf x),\mathbf y-R_{\mathbf\Phi}(\mathbf y)\rangle=\langle\mathbf x, \mathbf y-R_{\mathbf\Phi}(\mathbf y)\rangle\label{eq:prop1}
\end{equation}
where $R_{\mathbf\Phi}(\cdot)$ is the reconstruction of point with respect to subspace $\mathbf\Phi$.
\end{property}
\begin{IEEEproof}[Prop.\ref{prop1}]
Since $\mathbf y-R_{\mathbf\Phi}(\mathbf y)$ is orthogonal to subspace $\mathbf\Phi$ and $R_{\mathbf\Phi}(\mathbf x)$ lies in subspace $\mathbf\Phi$, hence $\langle R_{\mathbf\Phi}(\mathbf x), \mathbf y-R_{\mathbf\Phi}(\mathbf y)\rangle=0$ which is equivalent to Eq. \ref{eq:prop1}. 
\end{IEEEproof}
\begin{lemma}
\begin{equation}\label{theo1:eq1}
\langle\gamma_i^{(k)},\varepsilon_{k-1}(\mathbf{x})\rangle=\langle\mathbf{\psi}_i,\mathbf{x}-R_{\mathbf{\Phi}_{k-1}}(\mathbf{x})\rangle~.
\end{equation}
\end{lemma}
\begin{lemma}  The norm of reconstruction residue of basis $\psi_i$ with respect to subspace $R_{\mathbf{\Phi}_{k-1}}$ can be calculated recursively according to
 \begin{equation}\label{theo1:eq2}
\parallel\gamma_i^{(k)}\parallel^2=\parallel\gamma_i^{(k-1)}\parallel^2-\frac{|\langle\varphi_{k-1},\mathbf{\psi}_i\rangle|^2}{\parallel\varphi_{k-1}\parallel^2}
~.
\end{equation}
\label{theo1}
\end{lemma}
\begin{IEEEproof}See Appendix.\end{IEEEproof}

The denominator for each base vector
$\parallel\gamma_i^{(k)}\parallel^2$ can be easily updated in each
iteration, because the inner product
$\langle\varphi_k,\mathbf{\psi}_i\rangle$ can be quickly computed.

It is worth noting that reconstruction for any $\mathbf{x}$ (i.e.
$R_{\mathbf{\Phi}_k}(\mathbf{x})$) can be efficiently computed by
pre-computing
$\mathbf{\Phi}_k(\mathbf{\Phi}_k^\mathsf{T}\mathbf{\Phi}_k)^{-1}$. The
calculation of $\mathbf{\Phi}_k^\mathsf{T}\mathbf{x}$ is  the inner products
between $\mathbf{x}$ and the base vectors, which can be accomplished
in $O(k)$ time using integral image. Thus, computing the reconstruction $R_{\mathbf{\Phi}_k}(\mathbf{x})$ simply costs
$O(kWH)$ time, where $W,H$ are respectively the width and height of
the image template. As $\langle\varphi_k,\mathbf{x}\rangle$ and
$\parallel\mathbf{x}-R_{\mathbf{\Phi}_{k-1}}(\mathbf{x})\parallel^2$
can be pre-computed, the total computational complexity is $O(N_\psi
K(N_f+N_b))$ with $N_\psi$ the number of features in dictionary.

Below is the pseudo-code for D-OOMP where $\Sigma(\mathbf{x})$ represents the integral image of $\mathbf{x}$ and $\textsc{prod}(\psi_i,\Sigma(\mathbf{x}))$ represents the inner product between Haar-like feature $\psi_i$ and
an arbitrary vector $\mathbf{x}$ which is calculated using its integral image.

\newtheorem{algorithm}{Algorithm}

\noindent \textbf{Algorithm 1.} D-OOMP for Haar-like features

\begin{algorithmic}[1]
\STATE Initialize dictionary $\mathbf{\Psi}=[\psi_1,\psi_2,\ldots,\psi_{N_\psi}]$.
\STATE $denom(0,i)\gets \parallel\psi_i\parallel^2, \forall i\in [1,N_\psi]$
\FOR {$k = 1$ \TO $K$}
\FOR {$i=1$ \TO $N_\psi$}
\STATE $t\gets \textsc{prod}(\psi_i, \Sigma(\overline{\varphi_{k-1}}))$
\STATE $denom(k,i)\gets denom(k-1,i)-t^2$
\STATE $num\gets0$
\FOR {$j=1$ \TO $N_f$}
\STATE $num \gets num + \frac{1}{N_f}\textsc{prod}(\psi_i, \Sigma(\epsilon_{k-1}(\mathbf{f}_j)))$
\ENDFOR
\FOR {$j=1$ \TO $N_b$}
\STATE $num \gets num - \frac{\lambda}{N_b}\textsc{prod}(\psi_i, \Sigma(\epsilon_{k-1}(\mathbf{b}_j)))$
\ENDFOR
\STATE $\textrm{score}_i\gets {num}/{denom(k,i)}$
\ENDFOR
\STATE The $k$-th basis is $\psi_{opt}$ s.t. $opt=\arg\min_i{\textrm{score}_i}$.
\ENDFOR

\end{algorithmic}

\section{Faster Computation of D-OOMP }\label{sec:fast}
Although the recursive computation of $\|\gamma_i^{(k)}\|^2$ improves the efficiency of D-OOMP, the optimization process is still slow due to the huge number of features in the dictionary. Another reason is that the computation of scores is proportional to the number of samples $N_f+N_b$. In this section we develop two algorithms to significantly reduce the amount of computation. The first one is an exact algorithm called Iterative D-OOMP that reduces the redundant computation in each iteration of feature selection with a recursive formulation. The second is an approximate method named hierarchical D-OOMP that uses hierarchical search to reduce the search space. We show that combining the two methods can achieve significant computational savings.

\subsection{Iterative D-OOMP}
In the above implementation, the time complexity of maximizing Eq. \ref{eq:doomp} for each feature is $O(N_f+N_b)$.
Therefore, with the total number of foreground and background samples increasing, the computational load increases.
This computation bottleneck will limit the applications of DNBS. Thus, we design to compute the feature scores iteratively with an equivalent formulation
in which the foreground/background terms in Eq. \ref{dnbs:form} can be combined together and the time complexity will not be sensitive to the increasing of the example number.

To begin with, we denote $L_k(\psi_i)$ the item in Eq. \ref{eq:doomp} such that
\begin{equation}
\begin{split}
&L_k(\psi_i)=\\
&\frac{1}{N_f}\sum_{j=1}^{N_f}\frac{|\langle\gamma_i^{(k)},\varepsilon_{k-1}(\mathbf{f}_i)\rangle|^2}{\parallel\gamma_i^{(k)}\parallel^2}
-\frac{\lambda}{N_b}\sum_{j=1}^{N_b}\frac{|\langle\gamma_i^{(k)},\varepsilon_{k-1}(\mathbf{b}_i)\rangle|^2}{\parallel\gamma_i^{(k)}\parallel^2}
\end{split}
\end{equation}
The efficient computation of $L_k(\psi_i)$ plays a decisive role in speeding up the whole feature selection algorithm since it is exhaustively and repetitively calculated in each of the iterations. Through a series of equivalent transformations, we have the following proposition.
\begin{proposition}\label{theo:final}
During the selection procedure of the $k$-th basis, $L_k(\psi_i)$ can be calculated iteratively with known $L_{k-1}(\psi_i)$ such that
\begin{equation}\label{eq:theo:final}
\begin{split}
&L_k(\psi_i)=\\
&\frac{1}{d_i^{(k)}}\left[d_i^{(k-1)}L_{k-1}(\psi_i)-2\frac{\beta_i^{(k)}}{u_{k-1}}\langle\psi_i,\mathbf{I}_k\rangle+\left(\frac{\beta_i^{(k)}}{u_{k-1}}\right)^2S_k\right]
\end{split}
\end{equation}
where $d_i^{(k)}=\parallel\gamma_i^{(k)}\parallel^2$, $u_k=\parallel\varphi_{k}\parallel^2$ and
$\beta_i^{(k)}=\langle\psi_i,\varphi_{k-1}\rangle$.
\begin{equation}\label{theo:Ik}
\mathbf{I}_k=\frac{1}{N_f}\sum_{j=1}^{N_f}\eta_k(\mathbf{f}_j)-\frac{\lambda}{N_b}\sum_{j=1}^{N_b}\eta_k(\mathbf{b}_j)
\end{equation}
where $\eta_k(\mathbf{x})=\langle\varphi_{k-1},\mathbf{x}\rangle\varepsilon_{k-2}(\mathbf{x})$
\begin{equation}\label{theo:Sk}
S_k=\frac{1}{N_f}\sum_{j=1}^{N_f}\alpha_k^2(\mathbf{f}_j)-\frac{\lambda}{N_b}\sum_{j=1}^{N_b}\alpha_k^2(\mathbf{b}_j)
\end{equation}
with $\alpha_k(\mathbf{x})=\langle\varphi_{k-1},\mathbf{x}\rangle$~.

\label{theo1:eq}
\end{proposition}
\begin{IEEEproof}
See Appendix.
\end{IEEEproof}

It can be found from the above proposition that neither $\mathbf{I}_k$ or $S_k$ are related to
$\psi_i$, which indicates that they are same to each $\psi_i$ and can be pre-computed before the
main iteration of feature scoring. Then the computation of each feature score can be accomplished by
with only two inner product calculations based on integral images and several multiplications.
However, Eq. \ref{eq:theo:final} only applies for situations when $k > 1$. Thus, the
first binary base still has to be selected by the brute-force search,
which theoretically costs $N_\psi(N_f+N_b)$ operations where $N_\psi$ is the dictionary size. Let $K$ be the expected number of features,
$(W,H)$ the size of template and $N_f,N_b$ the numbers of foreground/background samples. Therefore, the time complexity of our approach achieves
\begin{equation}\nonumber
\begin{split}
&O(N_\psi(N_f+N_b)+KN_\psi+KWH(N_f+N_b)+K^2WH)\\
&=O((N_\psi+KWH)(K+N_f+N_b))\\
&=O((N_\psi+KN_\text{pix})(N_s+K))
\end{split}
\end{equation}
where $N_\text{pix}=WH$ is the number of pixels in template and $N_s = N_f + N_b$ is the total number of samples.

\noindent \textbf{Algorithm 2.} Iterative D-OOMP for Haar-like features

\begin{algorithmic}[1]
 \STATE Initialize dictionary $\mathbf{\Psi}=[\psi_1,\psi_2,\ldots,\psi_{N_\psi}]$.
\STATE Select the first feature $\varphi_1$ and initialize data.
\FOR {$k = 2$ \TO $K$}
\STATE Pre-compute $\mathbf{I}_k$ according to Eq. \ref{theo:Ik}.
\STATE Pre-compute $S_k$ according to Eq. \ref{theo:Sk}.
\FOR {$i=1$ \TO $N_\psi$}
\STATE Calculate $\beta_i^{(k)}\gets \textsc{prod}(\psi_i, \Sigma(\overline{\varphi_{k-1}}))$.
\STATE Calculate $d_i^{(k)}$:$\|\gamma_i^{(k)}\|^2\gets\|\gamma_i^{(k-1)}\|^2-(\beta_i^{(k)})^2$
\STATE Calculate $L_k(\psi_i)$ according to Eq. \ref{eq:theo:final}.
\ENDFOR
\STATE The $k$-th basis is $\psi_{opt}$ s.t. $opt=\arg\min_i{L_k(\psi_i)}$.
\ENDFOR
\end{algorithmic}

The iterative approach is theoretically an equivalent formulation of the
original D-NBS, and thus it will not incur any additional error to
the results. The most repetitive items are pre-computed to avoid redundant computation. Furthermore, the efficiency of the iterative approach stays
much more stable as the number of samples increases, which is more suitable for those applications having
a large number of training images.

\begin{figure}[!htb]
\centering
\subfigure[DNBS]{
\includegraphics[width=.17\textwidth]{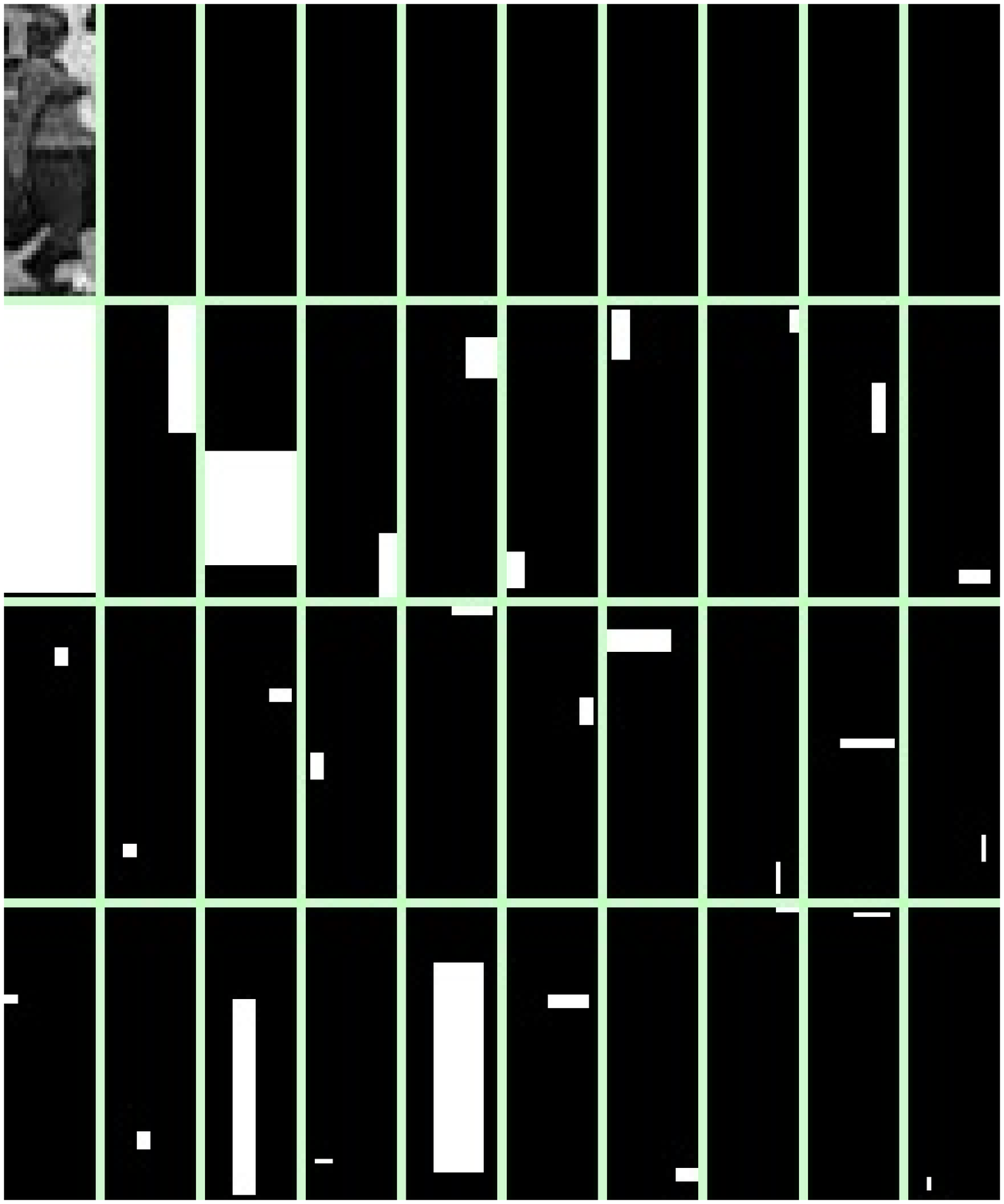}
}\hskip 5ex
\subfigure[NBS]{
\includegraphics[width=.17\textwidth]{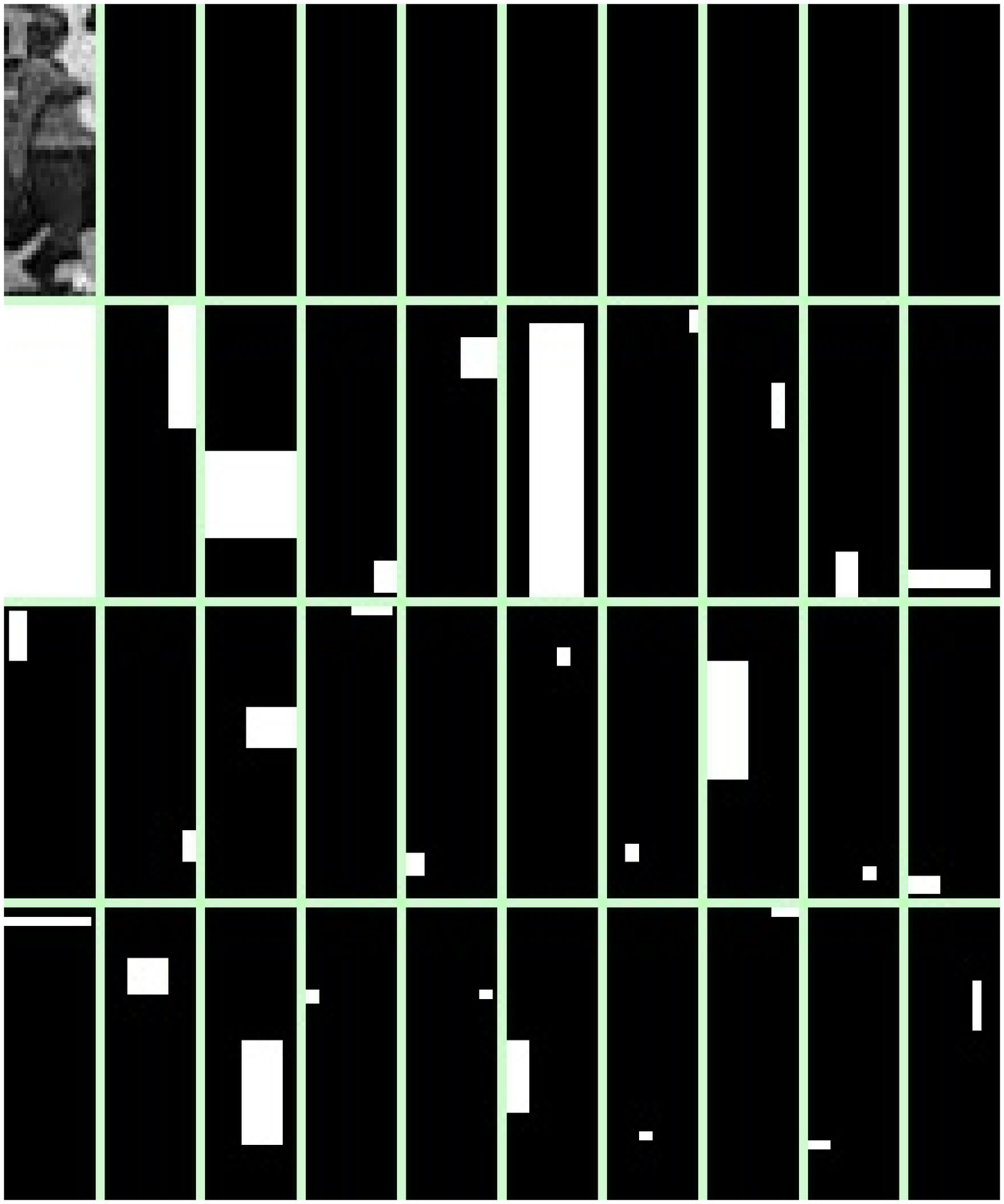}
}
   \caption{Top 30 features selected using Discriminative NBS (left) and the original NBS (right) for an image. The two feature sets are in general similar to each other while the differences between the two feature sets are due to the negative samples integrated into DNBS.}\label{fig:example_dnbs}
\label{arg}
\end{figure}

An example image and the selected Haar-like features using
Discriminative NBS are shown on the left of Figure
\ref{fig:example_dnbs}. It is compared with the results selected
using the original NBS shown on the right.

\subsection{Hierarchical D-OOMP}
As can be observed, the major computation cost for D-OOMP lies in searching all the features in the dictionary. Therefore, one natural way to speed up is to reduce the dictionary size. In this section, we propose a hierarchical searching approach to reduce the search space. The features in the dictionary is first grouped into clusters using a fast iterative clustering method. This forms a two level hierarchy with the first one as the cluster centers and the second as all the rest in the same cluster. During the search procedure, a Haar-like feature is first compared with each of the cluster centers so that the ones that are far away from the candidate feature can be easily rejected.

\subsubsection{Dictionary Clustering}
\begin{definition}[$\mu$-near basis set for a single feature]
For Haar-like basis $\varphi_i$, we define its $\mu$-near basis set to be $\mathcal{N}(\varphi_i,\mu)=\{\varphi_j|\langle\varphi_j,\varphi_i\rangle\ge\mu\}$.
\end{definition}

\begin{definition}[$\mu$-near basis set for a set of features]
 For a set of Haar-like basis $\Phi=\{\varphi_i\}$, we define its $\mu$-near basis set to be $\mathcal{N}(\Phi,\mu)=\cup_i{\mathcal{N}(\varphi_i,\mu)}$.
\end{definition}

All features in the dictionary are grouped into clusters such that
any feature has inner product larger than or equal to $\mu$ with the cluster center
(i.e. $\mu$-near basis set). The following three steps are iterated until all features have been assigned.
\begin{enumerate}
    \item Randomly select cluster center $c_i$ in the remaining feature set $\mathcal{F}$ ;
    \item Bundle features in $\mathcal{N}(c_i,\mu)$ to cluster $C_i$.
    \item Remove new cluster features: $\mathcal{F} = \mathcal{F} \setminus C_i$ .
\end{enumerate}
Afterwards, the dictionary $\mathbf{\Psi}$ is divided into groups of features $\mathbf{C}=\{C_1,C_2,\ldots\}$.

\subsubsection{Efficient Dictionary Clustering}
Observing that in the feature clustering process, the computation of $\mathcal{N}(c_i,\mu)$ is
the most expensive operation. We describe a fast $\mu$-near basis set retrieval method that leverages the special structural property of Haar-like box features.

For a cluster center $\phi$ and any feature $\psi$, the inner product is
\begin{equation}\label{1}
\langle \phi, \psi \rangle = \frac{\text{CommonArea}(\phi,\psi)}{\sqrt{\text{Area}(\phi)}\cdot\sqrt{\text{Area}{(\psi)}}}
\end{equation}
where $\text{CommonArea}(*,*)$ is the common area between the two rectangle features and $\text{Area}(*)$ denotes the area of a rectangle feature. In each iteration of the clustering algorithm, the center feature $\phi$ is selected and the remaining feature set is searched to select $\psi$'s that satisfy the inequality
\begin{equation}\label{2}
\langle \phi, \psi \rangle\ge \mu~.
\end{equation}
Integrating Eq.~\ref{1} and Eq.~\ref{2}, we get
\begin{equation}\label{3}
\frac{\text{CommonArea}(\phi,\psi)}{\sqrt{\text{Area}(\phi)}\cdot\sqrt{\text{Area}{(\psi)}}}\ge\mu~.
\end{equation}
Let $w_*$ and $h_*$ be the width and height of the non-zero rectangle of Haar-like feature $*$, then $\text{Area}(*)=w_*h_*$. The common area must be included in each of the two rectangle regions. A direct way is to search the bounding coordinates of feature $\psi$ and to calculate their intersections. However, this method would be too much expensive. We instead search the bounding coordinates of the common area and infer feature $\psi$ from the position of this common area. We suppose the common rectangle is of size $(w_\cap,h_\cap)$ and the extension from common rectangle to feature $\psi$ is $(l,r,t,b)$ indicating the left, right, top, and bottom margins respectively. Considering the fact that the common area between two rectangles is always a rectangle, we know that feature $\psi$ is of size $(w_\psi, h_\psi)=(w_\cap+l+r,h_\cap+t+b)$. Therefore, Eq.~\ref{3} can be re-written as
\begin{equation}
\frac{w_\cap h_\cap}{\sqrt{w_\phi h_\phi}\sqrt{(w_\cap+l+r)(h_\cap+t+b)}}\ge\mu
\end{equation}
and further simplified to
\begin{equation}
(w_\cap+l+r)(h_\cap+t+b)\le\frac{w^2_\cap h^2_\cap}{\mu^2w_\phi h_\phi}=A_\text{sup}~.
\end{equation}
Intuitively, in the case that the common rectangle is completely included in the rectangle $\phi$ (no edge overlapping), it is certain that $\psi$ is the same as the common rectangle. Thus, there should be limitations on the range of $(l,r,t,b)$. Here, $(x_*,y_*)$ is the coordinate of the top-left pixel of rectangle $*$.
\begin{equation}\nonumber
0\le l\le
\begin{cases}
\frac{A_\text{sup}}{h_\cap}-w_\cap,&x_\cap=x_\phi\\
0,&x_\cap\not=x_\phi
\end{cases}
\end{equation}
\begin{equation}\nonumber
0\le r\le
\begin{cases}
\frac{A_\text{sup}}{h_\cap}-w_\cap-l,&x_\cap+w_\cap=x_\phi+w_\phi\\
0,&x_\cap+w_\cap\not=x_\phi+w_\phi
\end{cases}
\end{equation}
\begin{equation}\nonumber
0\le t\le
\begin{cases}
\frac{A_\text{sup}}{(w_\cap-l-r)}-h_\cap,&y_\cap=y_\phi\\
0,&y_\cap\not=y_\phi
\end{cases}
\end{equation}
\begin{equation}\nonumber
0\le b\le
\begin{cases}
\frac{A_\text{sup}}{(w_\cap-l-r)}-h_\cap-t,&y_\cap+h_\cap=y_\phi+h_\phi\\
0,&y_\cap+h_\cap\not=y_\phi+h_\phi
\end{cases}
\end{equation}
Moreover, the size of common rectangle is limited to
\begin{equation}
w_\cap h_\cap\ge \mu^2w_\phi h_\phi~.
\end{equation}
Finally, constrained search of $(x_\cap, y_\cap, w_\cap, h_\cap, l, r, t, b)$ leads to a fast implementation of dictionary clustering.

With dictionary pre-clustered, each iteration of the Discriminative OOMP can be performed hierarchically. The cluster centers are examined first, only those clusters that are close enough are further searched. This approximate solution can significantly reduce the computation load required with minimal accuracy decreasing using carefully tuned parameter settings. The hierarchical D-OOMP algorithm is given as follows.

\noindent \textbf{Algorithm 3.} Hierarchical D-OOMP

\begin{algorithmic}[1]
 \STATE Initialize dictionary $\mathbf{\Psi}=[\psi_1,\psi_2,\ldots,\psi_{N_\psi}]$.
\STATE Cluster features with center index $\mathbf{C}=\{c_1, c_2,\ldots\}$ in accord with the given $\mu$.
\STATE Select the first feature $\varphi_1$ and initialize data.
\FOR {$k = 2$ \TO $K$}
\STATE Pre-compute $\mathbf{I}_k$ according to Eq.~\ref{theo:Ik}.
\STATE Pre-compute $S_k$ according to Eq.~\ref{theo:Sk}.
\FOR {$i=1$ \TO $|\mathbf{C}|$}
\STATE Calculate $\beta_{c_i}^{(k)}\gets \textsc{prod}(\psi_{c_i}, \Sigma(\overline{\varphi_{k-1}}))$.
\STATE Calculate $d_{c_i}^{(k)}$:$\|\gamma_{c_i}^{(k)}\|^2\gets\|\gamma_{c_i}^{(k-1)}\|^2-(\beta_{c_i}^{(k)})^2$
\STATE Calculate $L_k(\psi_{c_i})$ according to Eq.~\ref{eq:theo:final}.
\ENDFOR
\STATE Get the optimal index: $opt=\arg\max_{c_i}{L_k(\psi_{c_i})}$
\FOR {$i=1$ \TO $|\mathbf{C}|$}
\IF {$L_k(\psi_{c_i})>L_k(\psi_{opt}) - ratio|L_k(\psi_{opt})|$}
\FOR {each feature $\psi_j$ s.t. $\langle\psi_j,\psi_{c_i}\rangle\ge\mu$}
\STATE Calculate $L_k(\psi_j)$ according to Eq.~\ref{eq:theo:final}.
\STATE Update $opt=j$ if $L_k(\psi_j) > L_k(\psi_{opt})$.
\ENDFOR
\ENDIF
\ENDFOR
\STATE The $k$-th basis is $\phi_k = \psi_{opt}$.
\ENDFOR
\end{algorithmic}

\begin{figure}[!htb]
\centering
\includegraphics[width=.4\textwidth]{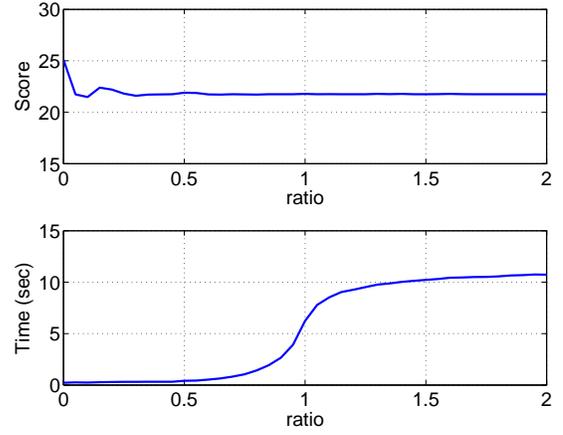}
\caption{Statistics on score and computational cost against the value of ratio in Hierarchical D-OOMP}
\label{fig1}
\end{figure}

In the $k$-th iteration of feature selection, all the cluster centers are scored. Supposing the maximum is $L_k^\textrm{(max)}=\max_{c_i}\{L_k(c_i)\}$, those groups whose central scores are bigger than $L_k^\textrm{(max)}-\textsc{ratio}\times|L_k^\textrm{(max)}|$ are further examined. It is obvious that this pruning operation will lose some precision when this threshold is limited. We aim to seek a balance between efficiency and accuracy of Hierarchical D-OOMP here.
The error score in Fig. \ref{fig1} is defined using the function in Eq. \ref{dnbs:form}.
According to Fig. \ref{fig1}, when \textsc{ratio} is between 0.3 and 0.6, the time consumption of the algorithm is relatively low (less than 1 seconds) while its accuracy is close to the original D-OOMP (when \textsc{ratio} is infinitely large). We empirically set it to 0.5 in experiments.

\subsection{Comparison of the Three Optimization Methods}
We compare the performance of the original D-OOMP,
Iterative D-OOMP and Hierarchical D-OOMP. As can be observed, there are several parameters that control the efficiency of D-OOMP, for example the number of bases selected and the number of samples used for training. The more features we need to select, the more time it takes. Also, in general, the more samples for training, the more computation it requires. In all
of the following experiments, all the image templates are all of size $50\times50$. All time statistics are calculated excluding pre-processing.

\begin{figure}[!htb]
\centering
\includegraphics[width=.4\textwidth]{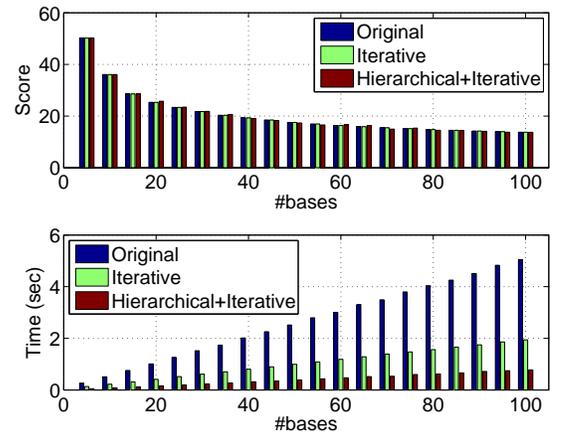}
\caption{Score and time against the number of bases, using 5 positive and 5 negative samples.}
\label{fig:nbase}
\end{figure}

In Fig. \ref{fig:nbase}, we show the relation between the number of bases and optimization score by varying the number of
bases from 1 to 100. As can be observed, the more bases, the better the solution is. The original D-OOMP and iterative D-OOMP have no difference in performance because the iterative D-OOMP is an equivalent transformation of the original D-OOMP. The hierarchical D-OOMP has slightly higher error because it yields an approximate solution.

\begin{figure}[!htb]
\centering
\includegraphics[width=.4\textwidth]{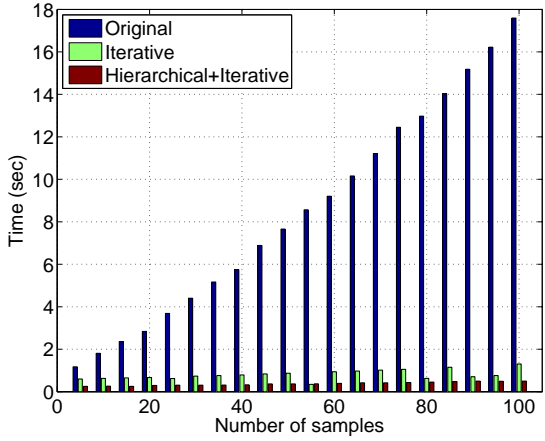}
\caption{Time consumption with respect to the number of background samples varying from 5 to 100.} \label{fig:nback}
\end{figure}

One of the major advantages of the two new efficient D-OOMP algorithms is that the computation
is not sensitive to the total number of foreground and
background samples. We here simply change the number of background
samples from 5 to 100 and see how the reconstruction error and
computation time change. The result is shown in Figure
\ref{fig:nback}. As can be observed, as the number of training samples increases, the computation cost for the original D-OOMP goes linearly while the computation for iterative D-OOMP and hierarchical D-OOMP remains stable.

\section{Tracking Using Discriminative NBS}\label{sec:track}
We apply the DNBS representation to visual object tracking. With the DNBS object representation, we locate object
position in the current frame through sum of squared difference
(SSD)-based matching. Using discriminative NBS, the object is first
compared with the possible locations in an region in the current frame around the predicted object
position. The one with the minimum
SSD value is chosen as the target object location. In order to accommodate
object appearance changes, the foreground and discriminative NBS are
automatically updated every few frames.

\subsection{Object Localization}\label{sec:ssd}
We use SSD to match the template, due to its high
efficiency of matching under the discriminative NBS representation.
In each frame $t$, we specify a rectangular region centered at the predicted
object location as the search window, in which the
templates are sequentially compared with the referenced foreground
$\mathbf{x}=R_{\mathbf{\Phi}_K^{(t)}}(\mathbf{f}_{\text{ref}}^{(t)})$.

Suppose that $\mathbf{x}$ is the object  and $\mathbf{y}$ is a
candidate object in the search window. The SSD between them
is,
\begin{equation}\label{ssd}
\text{SSD}(\mathbf{x},\mathbf{y})=\parallel\mathbf{x}-\mathbf{y}\parallel^2=\parallel\mathbf{x}\parallel^2+\parallel\mathbf{y}\parallel^2-2\langle\mathbf{x},\mathbf{y}\rangle
~,
\end{equation}
where $\parallel\cdot\parallel$ represents the $L^2$-norm and
$\langle\cdot,\cdot\rangle$ denotes the inner product. $\mathbf{x}$
is approximated by DNBS $\mathbf{\Phi}_K$ (i.e.
$R_{\mathbf{\Phi}_K^{(t)}}(\mathbf{f}_{\text{ref}}^{(t)})=\sum_{i=1}^K{c_i^{(t)}\phi_i^{(t)}}$)
, built using the approach in Section \ref{sec:fast}. Eq.~\ref{ssd}
is then transformed to
\begin{flalign}\label{ssd:nbs}
 \begin{split}
\text{S}&\text{SD}(\sum_{i=1}^K{c_i^{(t)}\phi_i^{(t)}}, \mathbf{y})\\
&=\parallel\sum_{i=1}^K{c_i^{(t)}\phi_i^{(t)}}\parallel^2+\parallel\mathbf{y}\parallel^2-2\sum_{i=1}^K{c_i^{(t)}}\langle\phi_i^{(t)}, \mathbf{y}\rangle ~.
 \end{split}
\end{flalign}
The first term is the same for all the candidate locations in the
current frame, while the second and third ones can be computed
rapidly using integral image. The online computational
complexity of Eq. \ref{ssd:nbs} is only $O(K)$, where $K$ is the
number of selected bases.

\subsection{Subspace Update}
Due to appearance changes of the object, the DNBS
built in the previous frame might be unsuitable for the current
frame. A strategy to dynamically update the subspace is necessary.
Here we update the subspace every 5 frames. Once a new subspace
needs to be computed, we first use the updated template and
background samples from the current frame to compute the DNBS again as Eq. \ref{dnbs:form}.

\subsubsection{Template Update}

The object template is also updated constantly to incorporate
appearance changes and the updated template serves as the new positive sample. According to
Eq. \ref{dnbs:form}, DNBS is then constructed to better represent the object using an updated set of samples. Intuitively, these sampled foregrounds
should recently appear, in order to more precisely describe the
current status of the object. Many previous efforts have been
devoted to template update (see \cite{pami03:update}). One natural
way is to choose the recent $N_f$ referenced foregrounds. Another
solution is to update the reference template in each frame, but this
may incur considerable error accumulation. Simply keeping it
unchanged is also problematic due to object appearance changes. A
feasible way is to update the foreground by combining the frames
using time-decayed coefficients. Here, we propose to update the
foreground reference for every $N_u$ frames,
\begin{equation}\nonumber
\mathbf{f}_{\text{ref}}^{(t)} =\left\{
\begin{array}{ll}
\mathbf{f}_0~&\quad t=0~\\
{\gamma}\mathbf{f}_{\text{ref}}^{(\lfloor (t-1)/N_u\rfloor
N_u)}+(1-\gamma)\mathbf{f}_t~&\quad \text{otherwise}~,
\end{array}\right.
\end{equation}
where $\mathbf{f}_0$ is the foreground  specified in the first frame
and $\mathbf{f}_t$ is the matched template at frame $t$. $\gamma$ is
the tradeoff, which is empirically set to $0.5$ in our experiments.
$\lfloor (t-1)/N_u\rfloor N_u$ is the frame at which the current
subspace is updated. $\mathbf{f}_{\text{ref}}^{(\lfloor
(t-1)/N_u\rfloor N_u)}$ is the object template at that frame. This
means we are updating the template periodically instead of at each
frame, which is more robust to tracking errors. This template
updating scheme is compared with other methods and the results are shown
in the experimental section.

\subsubsection{Background Sampling}

The background samples which closely resemble the reference
foreground often interfere with the stability and accuracy of
tracker. We sample the background templates which are similar to the
current reference object and take them as the negative data in
solving the DNBS. We compute a distance map in a
region around the object and those locations that are very similar
to the object are selected as the negative samples. This
process can be done efficiently because the SSD distance map
can be computed efficiently using Haar-like features and
integral images. Once the distance map is computed, the local minima locations are used to
select negative training examples by means of non-minimal suppression.

\section{Experiments}\label{sec:experiment}

The proposed approach is evaluated on a set of sequences extracted from public video datasets. These sequences are challenging because of their background clutter and camera motion. Some key parameters, such as $\lambda$ used in the DNBS formulation and $\mu$ used in hierarchical D-OOMP, are firstly discussed in this section. The qualitative
tracking results are shown afterwards. To demonstrate the
advantages of our approach and the benefit of the discriminative term in DNBS, we qualitatively compare our tracker with
 an NBS tracker which applies the original
NBS object representation. We show in this comparison that the discriminative terms in DNBS help increase the tracking accuracy. Quantitative evaluations are conducted by comparing the success rates of our tracker against several state-of-the-art trackers. In addition, we also provide a comprehensive comparison by employing the evaluation protocols proposed by \cite{WuLimYang13}. While achieving a relatively stable performance, our tracker is able to be processed in real-time.


\subsection{Parameter Selection}

Several parameters are used in the DNBS such as the trade-off $\lambda$ between foreground and background reconstruction errors. Intuitively, those parameters
can influence the accuracy of object
reconstruction and the tracking performance. So we perform experiments on these paramters and discuss the justification of
the selections.

The formulation of the DNBS balances the influence of
the foreground and background reconstruction terms with a
coefficient $\lambda$. Intuitively, it should be set to a small
value to ensure the accuracy of foreground representation. To find
the best value, we use several image sequences ({\em``Browse"}, {\em``Crosswalk"} and {\em``OccFemale"}) with ground-truth to quantitatively evaluate how this
parameter affects the tracking accuracy.  To generate more data for evaluation, we split each of the
sequences into multiple subsequences initialized at different frames.

\begin{figure}[!ht]
\centering
\subfigure[]{
\includegraphics[width=.24\textwidth]{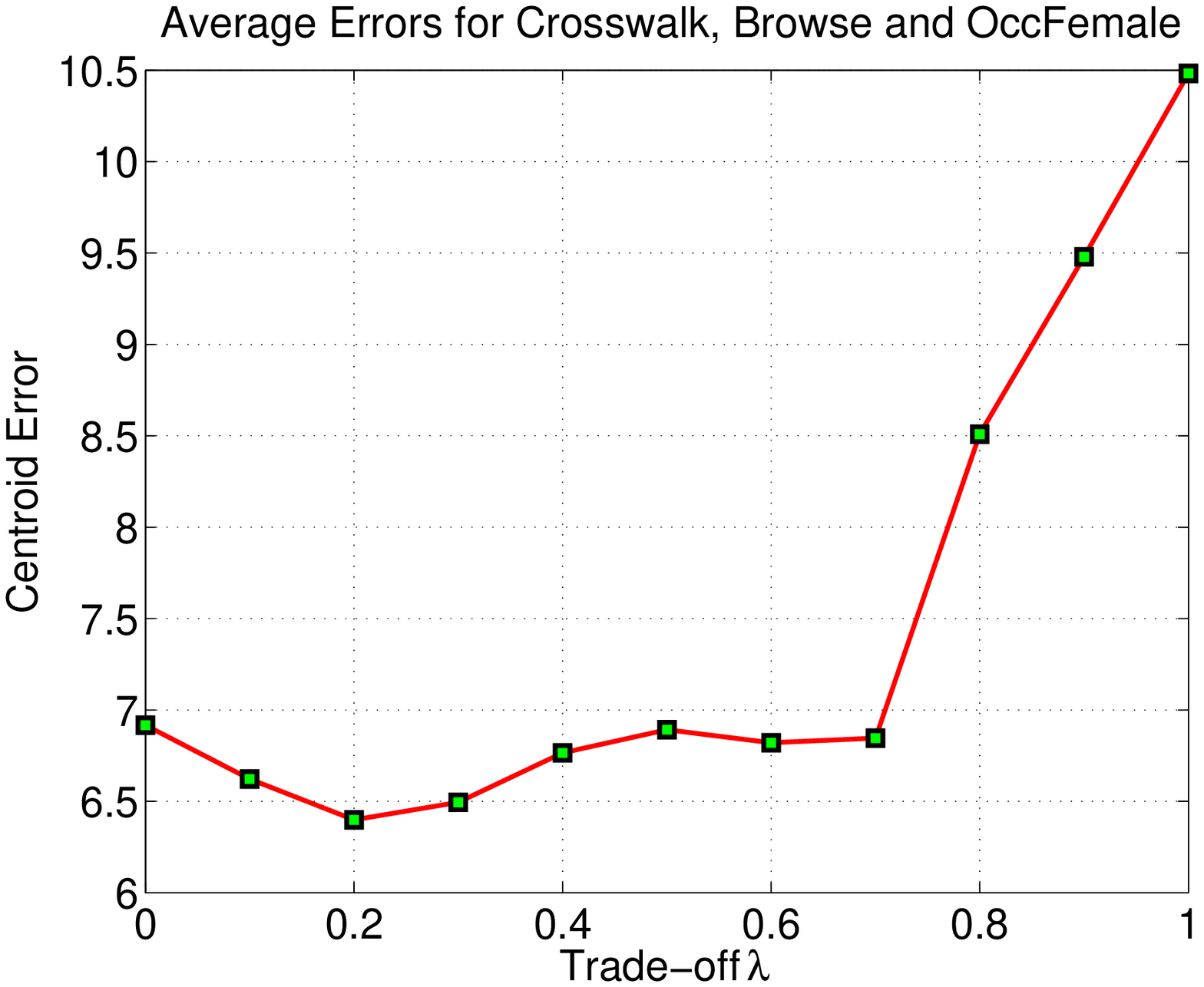}
\label{fig:param}}\subfigure[]{
\includegraphics[width=.24\textwidth]{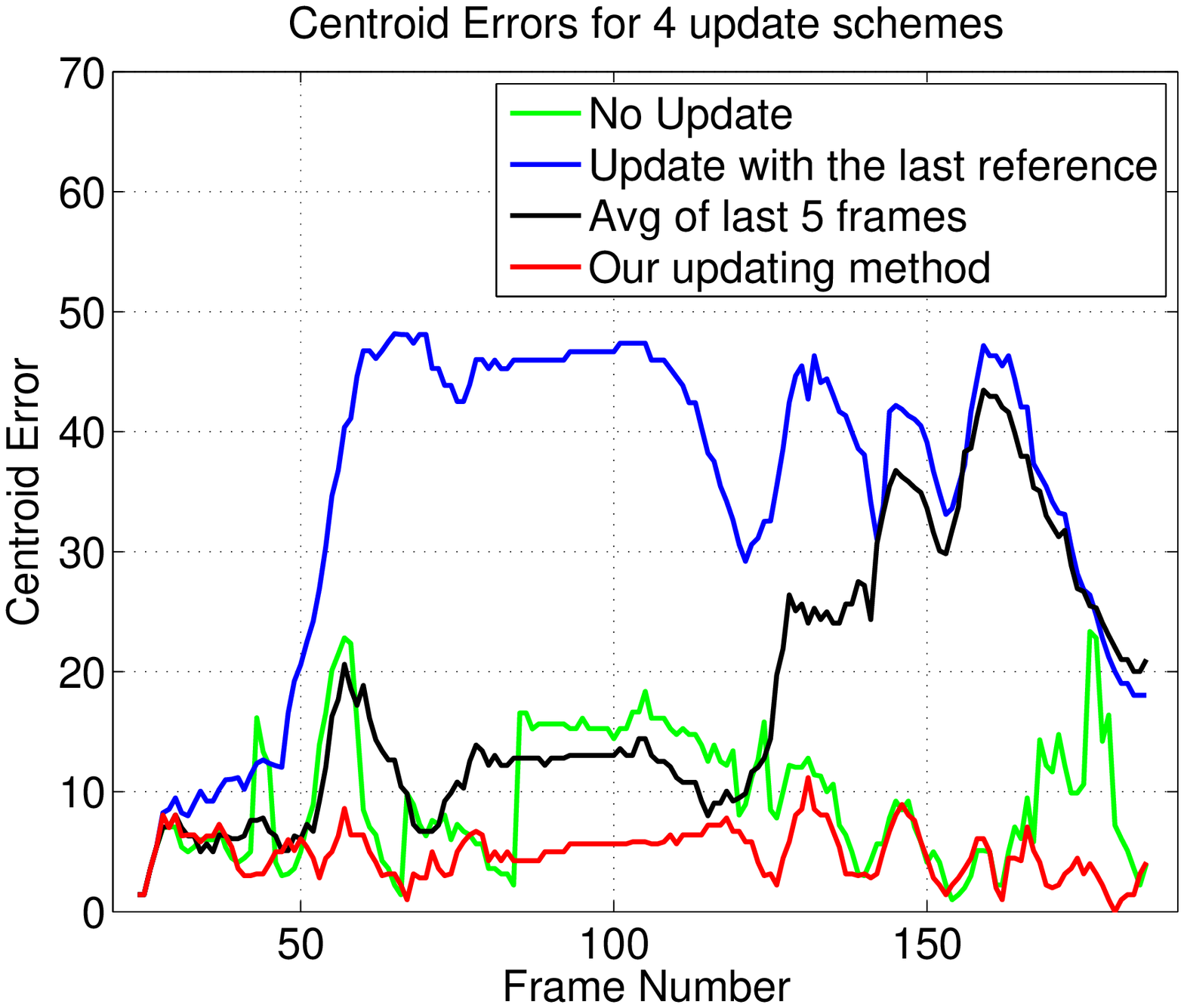}\label{fig:templateupdate}}
\caption{(a) The influence of $\lambda$ on the averaged tracking errors on multiple sequences. (b) Performance (centroid tracking errors) comparison among the four template updating schemes.}
\end{figure}

The tracking performance is
evaluated using the mean distance error between the tracked object location
and the groundtruth object center. 
Specifically, we initialize our tracker in each of the frame using the
groundtruth as the bounding box and record average tracking errors for the
subsequent 20 frames with different choices of the parameter $\lambda$. For each
sequence, errors of all the subsequences under the same $\lambda$ are
averaged and plotted in Fig. \ref{fig:param}. As is observed, the centroid error is relatively more
stable and smaller when $\lambda$ is set to 0.25.

Another parameter for DNBS is the number of bases $K$
used. The selection of this parameter depends on image content. In
general, the more features we use, the more accuracy DNBS is able to reconstruct the object. However, more features bring more computational costs. As a tradeoff, we set $K=30$. We empirically set the number of
foreground templates $N_f$ to 3 and that of background ones $N_b$ to $3$.
These parameters are fixed for all the experiments.

We also conduct experiments to show the effectiveness of our
template updating scheme. Here, we review several template updating
methods mentioned above by comparing their tracking errors of video
sequence {\em Browse}. These updating schemes include: 1) updating
the current template with the previous one; 2) updating the current
template with an average of previous 5 frames and our updating
method. All of the schemes are initialized with the same bounding
box at the first frame and the error of object center is computed
with respect to the groundtruth. Fig. \ref{fig:templateupdate} shows
that the time-decaying approach is more robust and stable.

\begin{figure}[!htb]
  \centering
  \includegraphics[width=.4\textwidth]{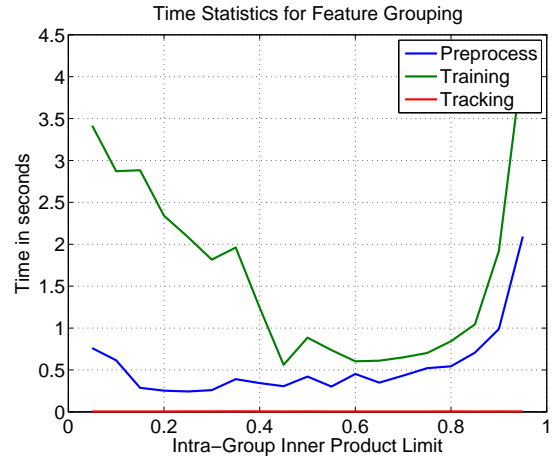}
  \caption{Time statistics on building dictionary hierarchy (preprocessing), subspace feature selection (training) and object localization (tracking per frame) for Sequence {\em Crosswalk} with respect to $\mu$.}
  \label{fig:timestats}
\end{figure}

\begin{figure*}
\def\imheight{3cm}
\centering
\subfigure[Frame 1]{
\includegraphics[height=\imheight]{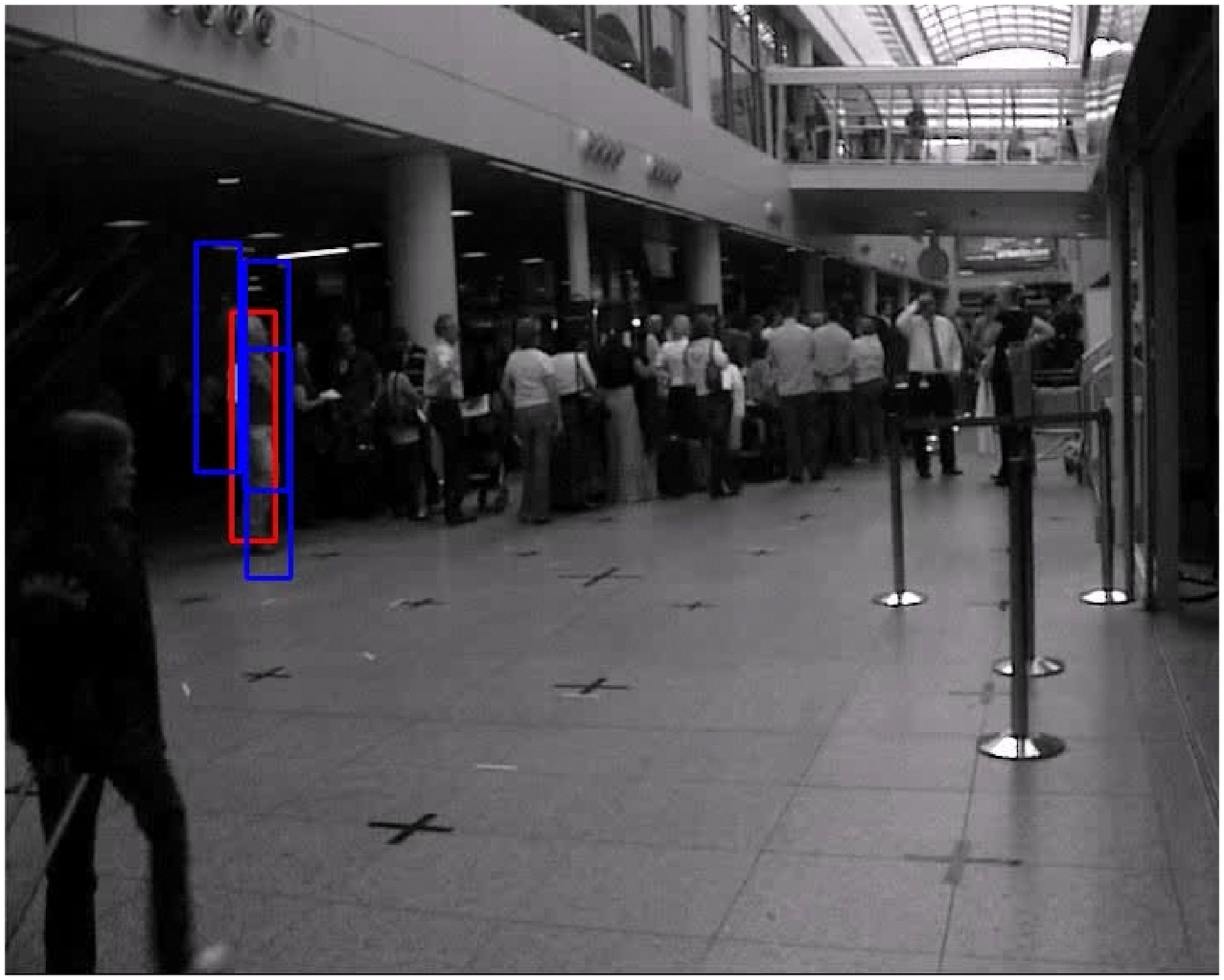}
}
\subfigure[Frame 18]{
\includegraphics[height=\imheight]{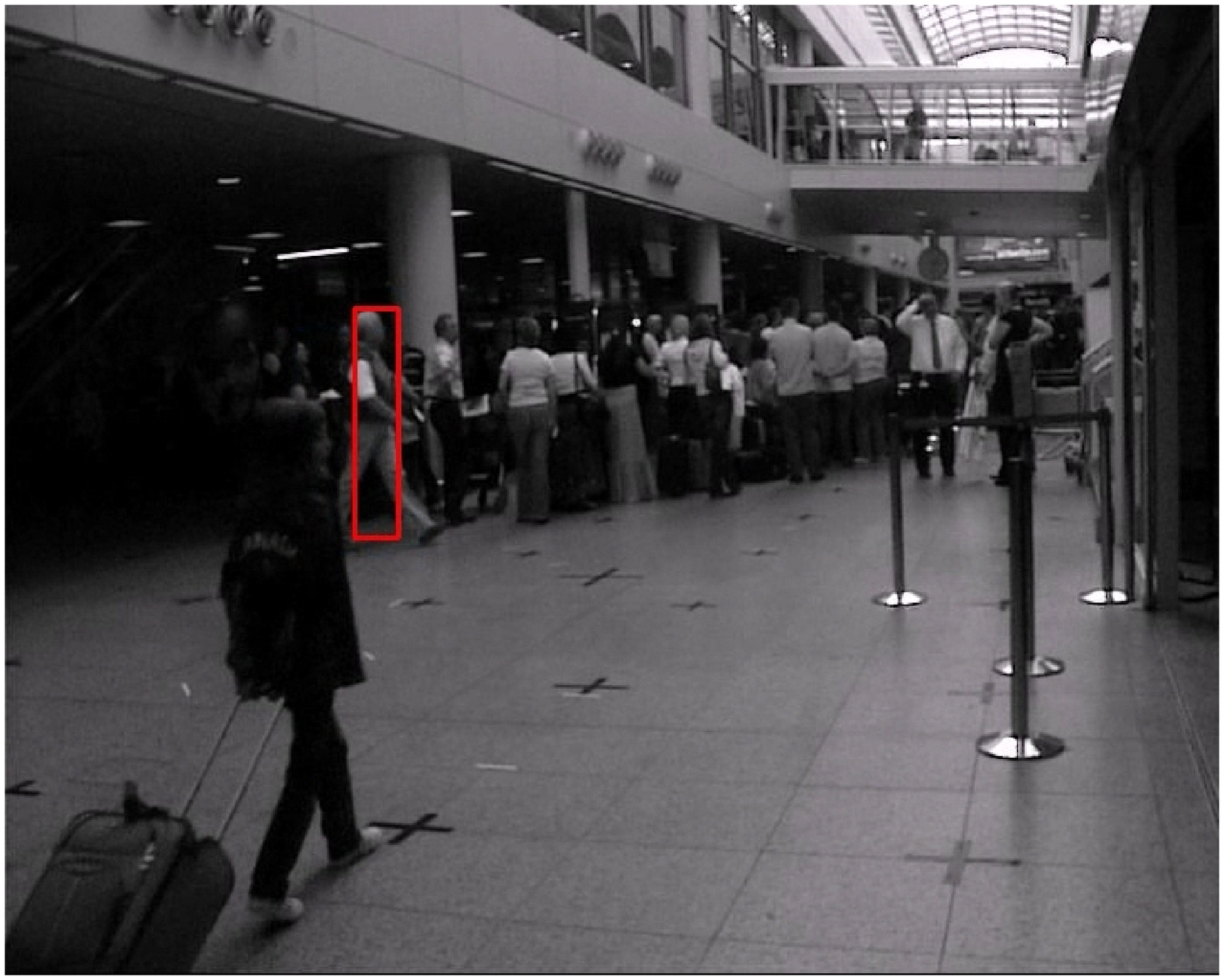}
}
\subfigure[Frame 46]{
\includegraphics[height=\imheight]{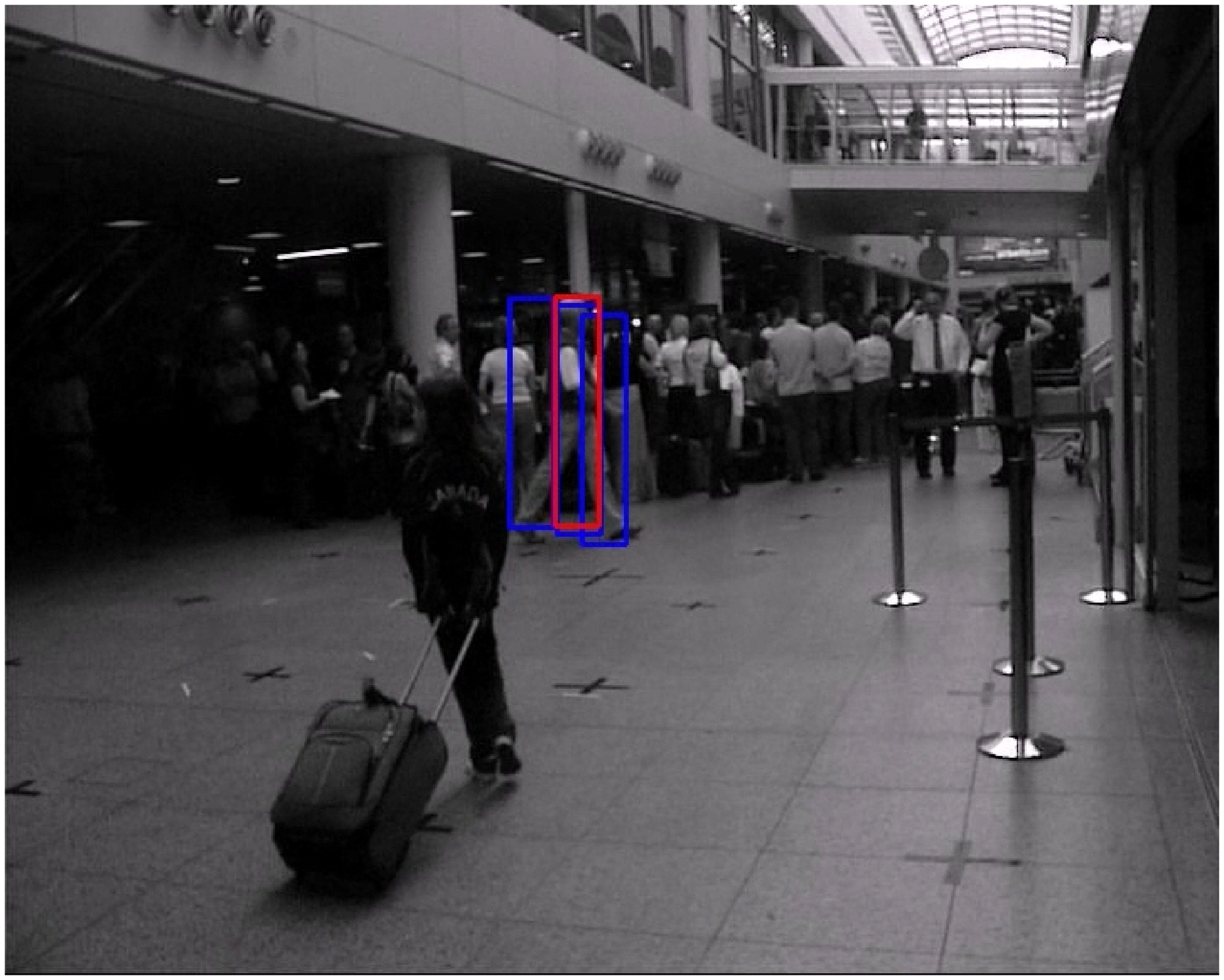}
}\\
\subfigure[Frame 57]{
\includegraphics[height=\imheight]{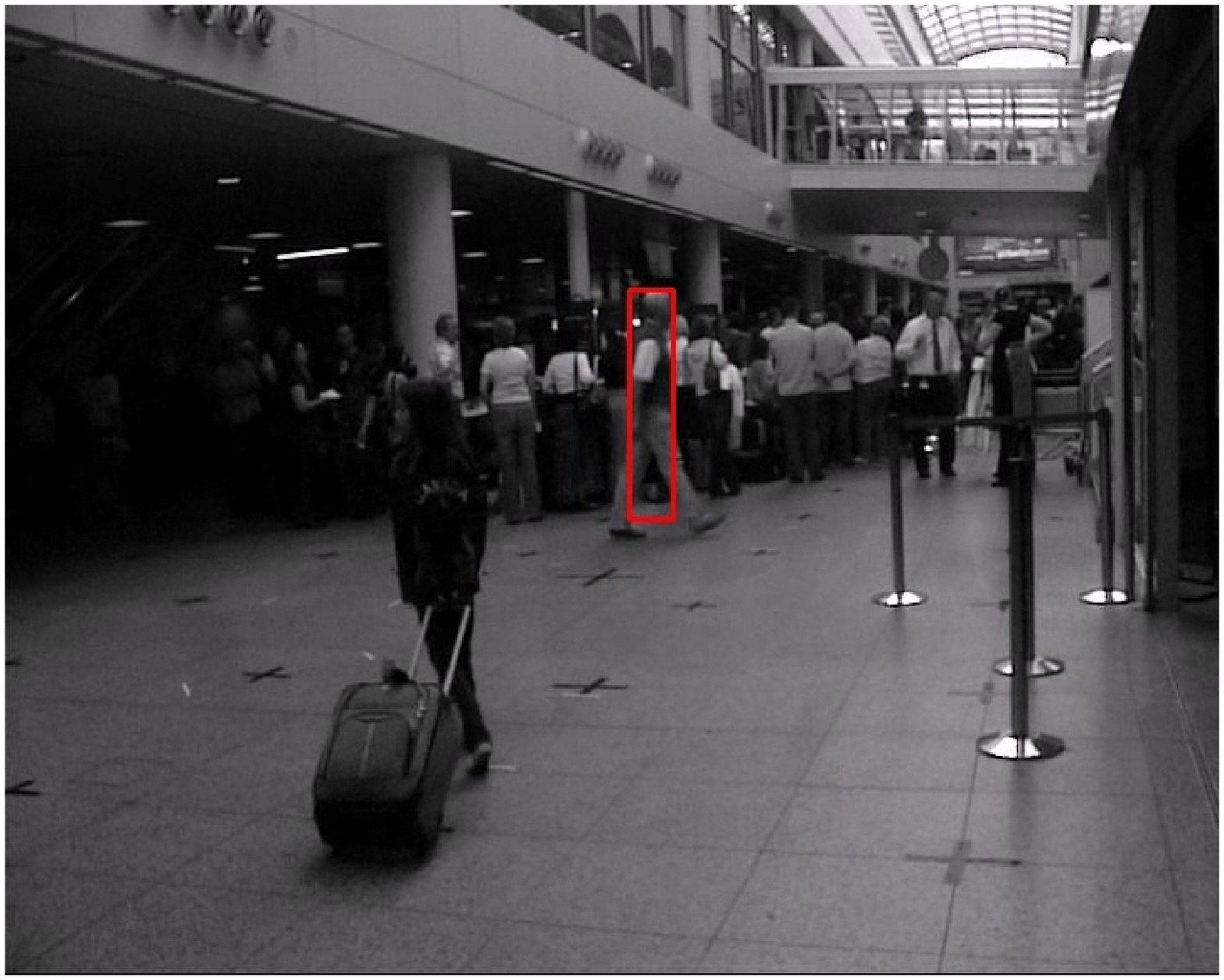}
}
\subfigure[Frame 76]{
\includegraphics[height=\imheight]{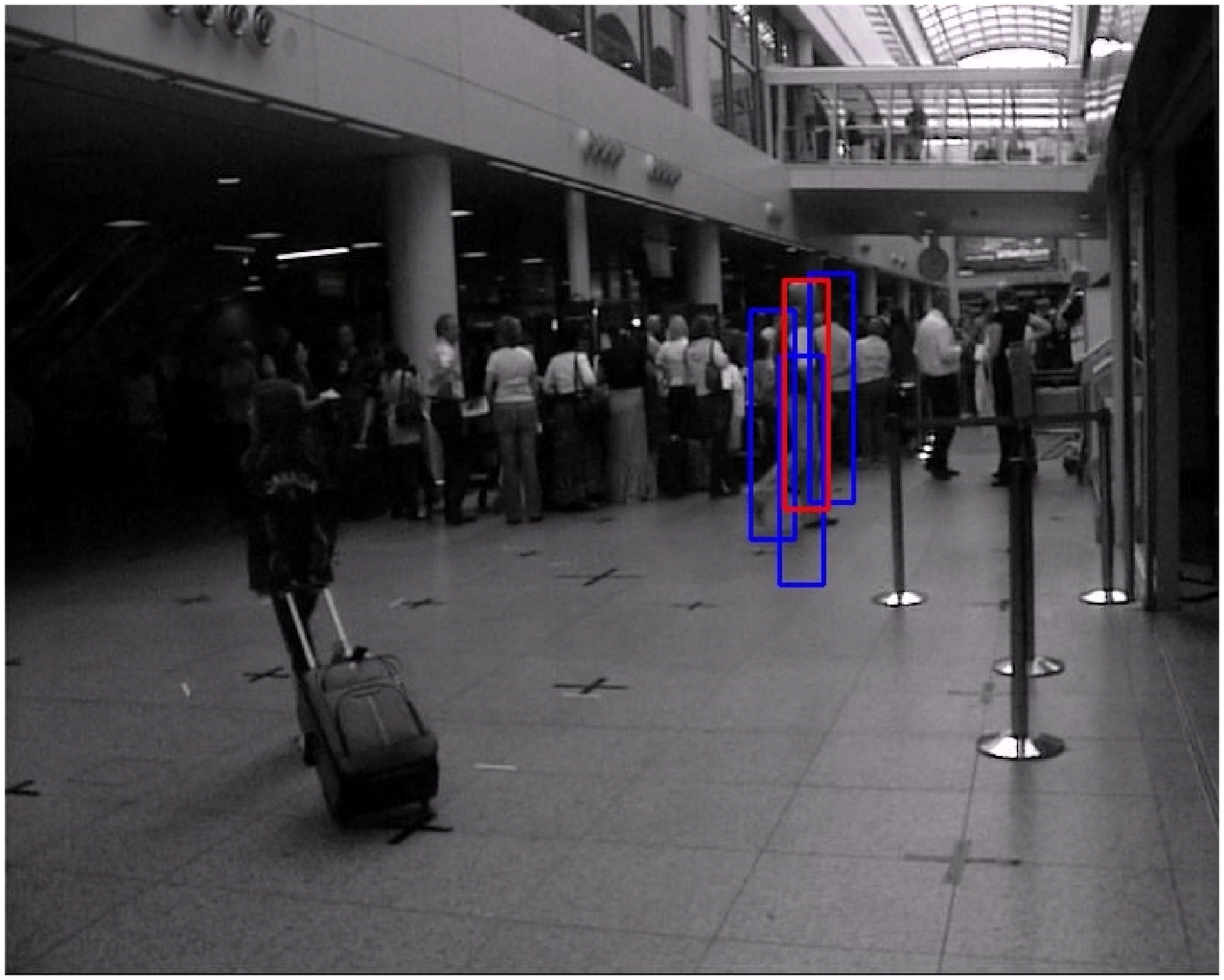}
}
\subfigure[Frame 89]{
\includegraphics[height=\imheight]{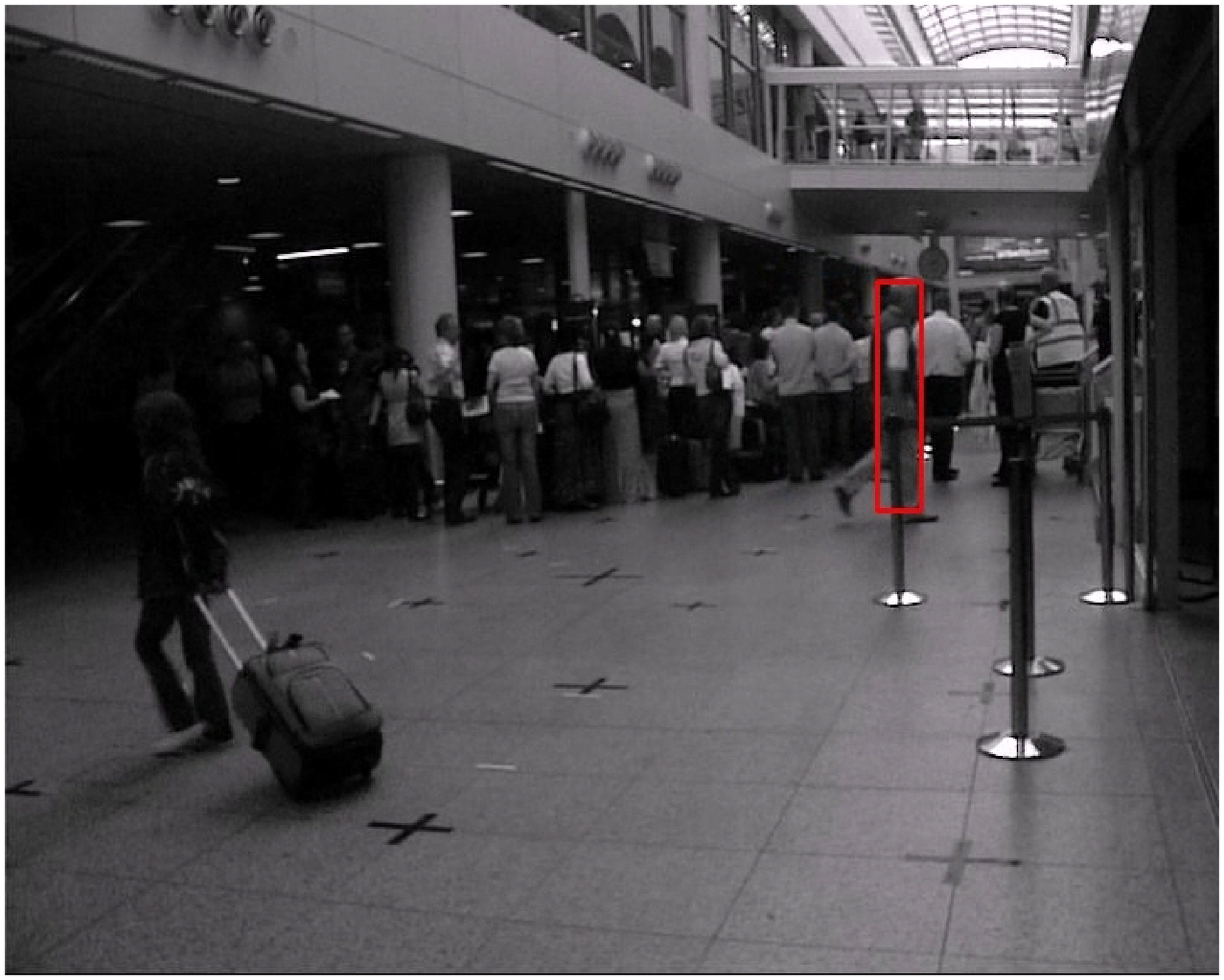}
}
\caption{Sequence {\em Crowd}: The frames 1, 18, 46, 57, 76 and 89 are shown. The red boxes are tracked objects using DNBS, the green boxes are tracking results using NBS and the blue boxes
in some of the frames are sampled backgrounds for subspace update in DNBS.}
\label{fig:crowd}
\end{figure*}

\begin{figure*}
\def\imheight{2.7cm}
\centering
\subfigure[Frame 1]{
\includegraphics[height=\imheight]{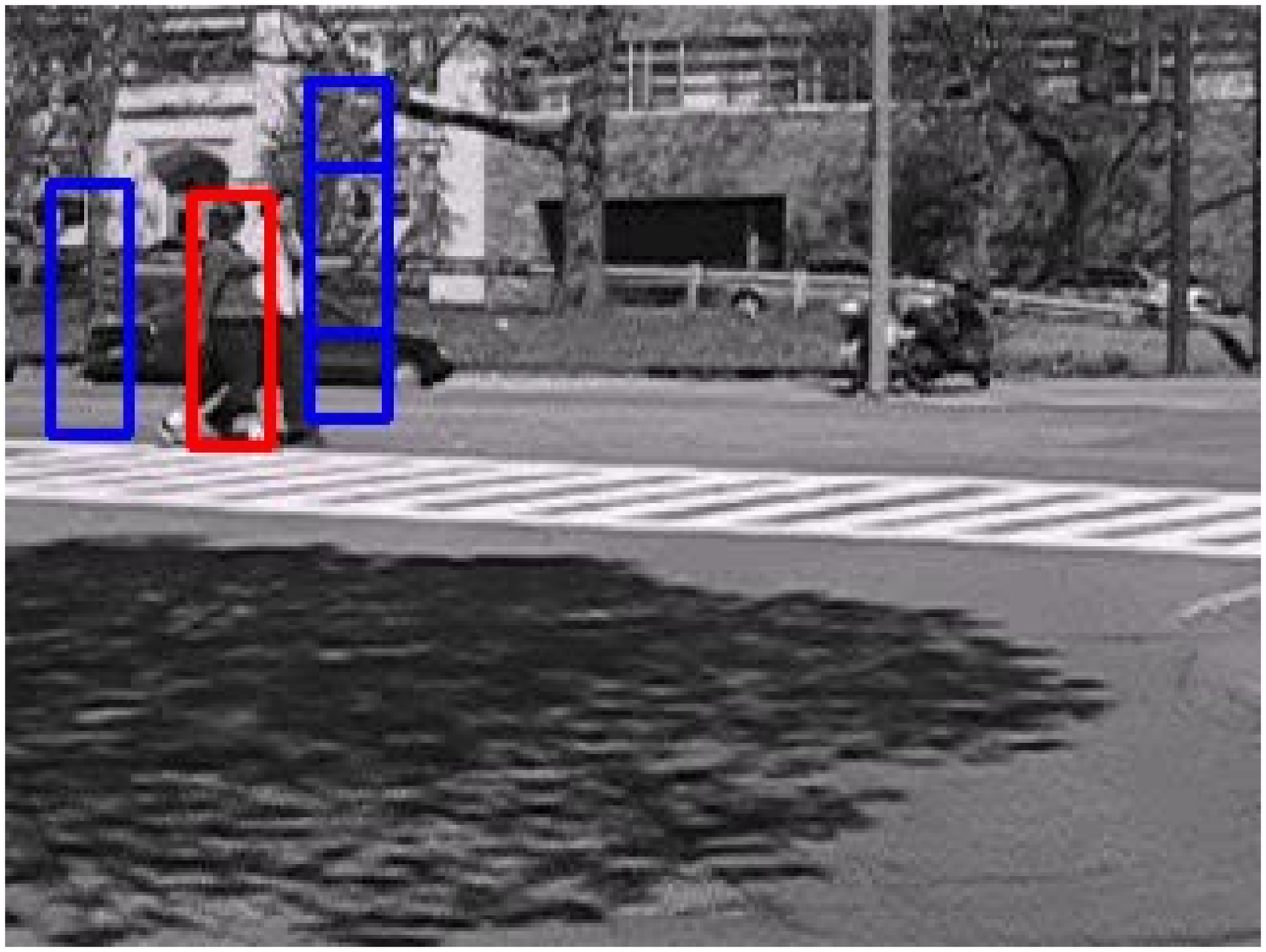}
}
\subfigure[Frame 17]{
\includegraphics[height=\imheight]{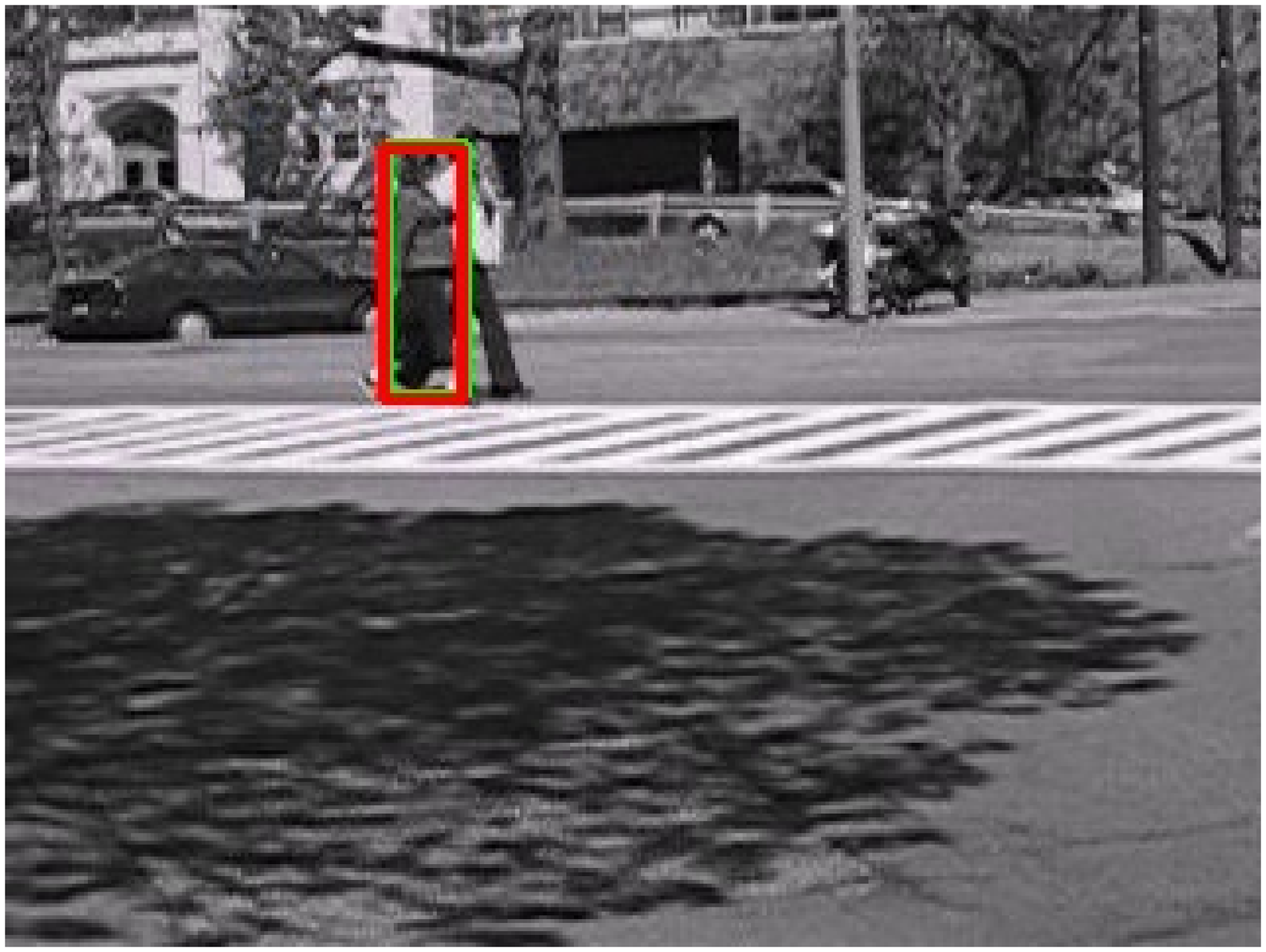}
}
\subfigure[Frame 51]{
\includegraphics[height=\imheight]{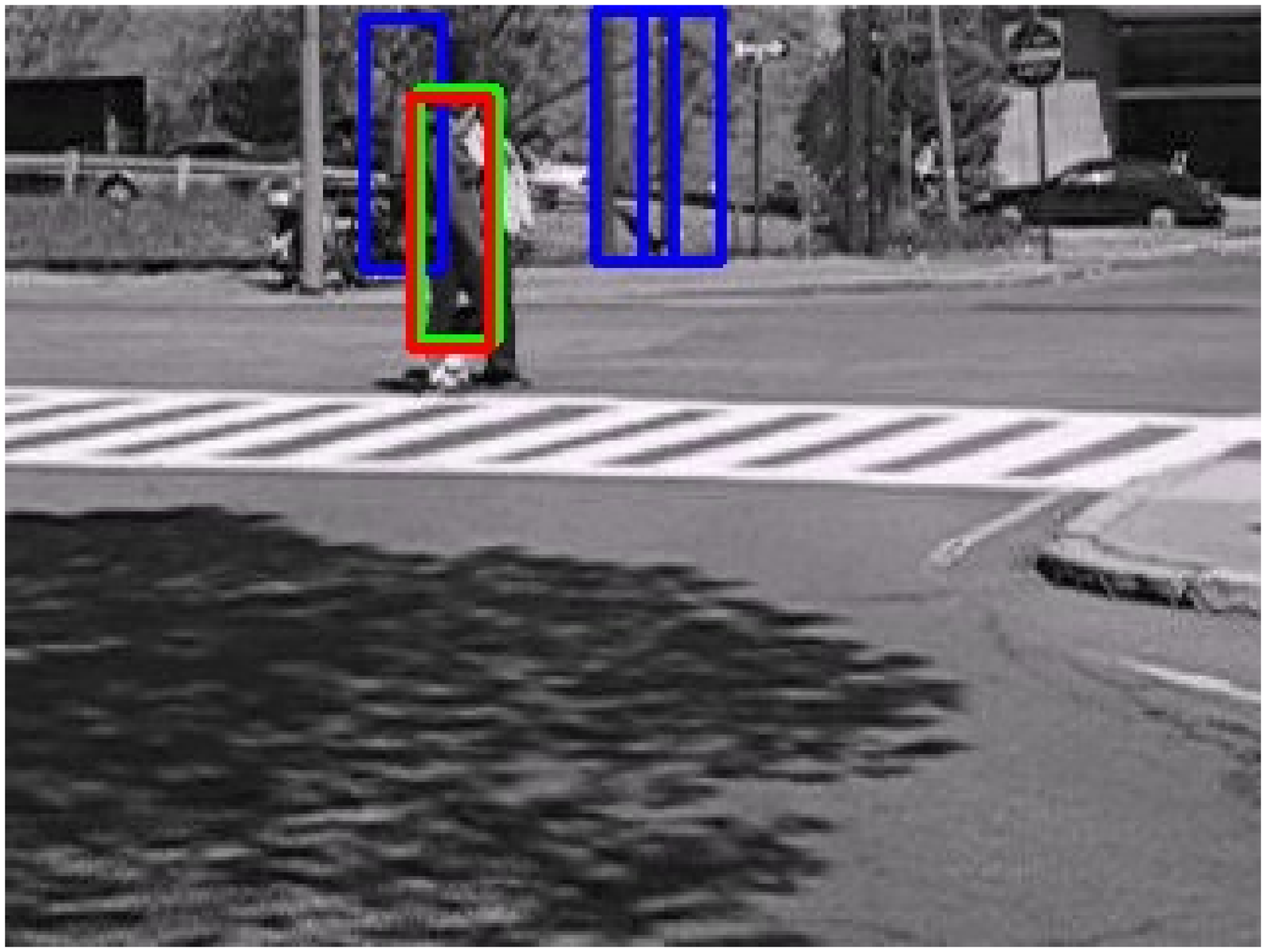}
}\\
\subfigure[Frame 75]{
\includegraphics[height=\imheight]{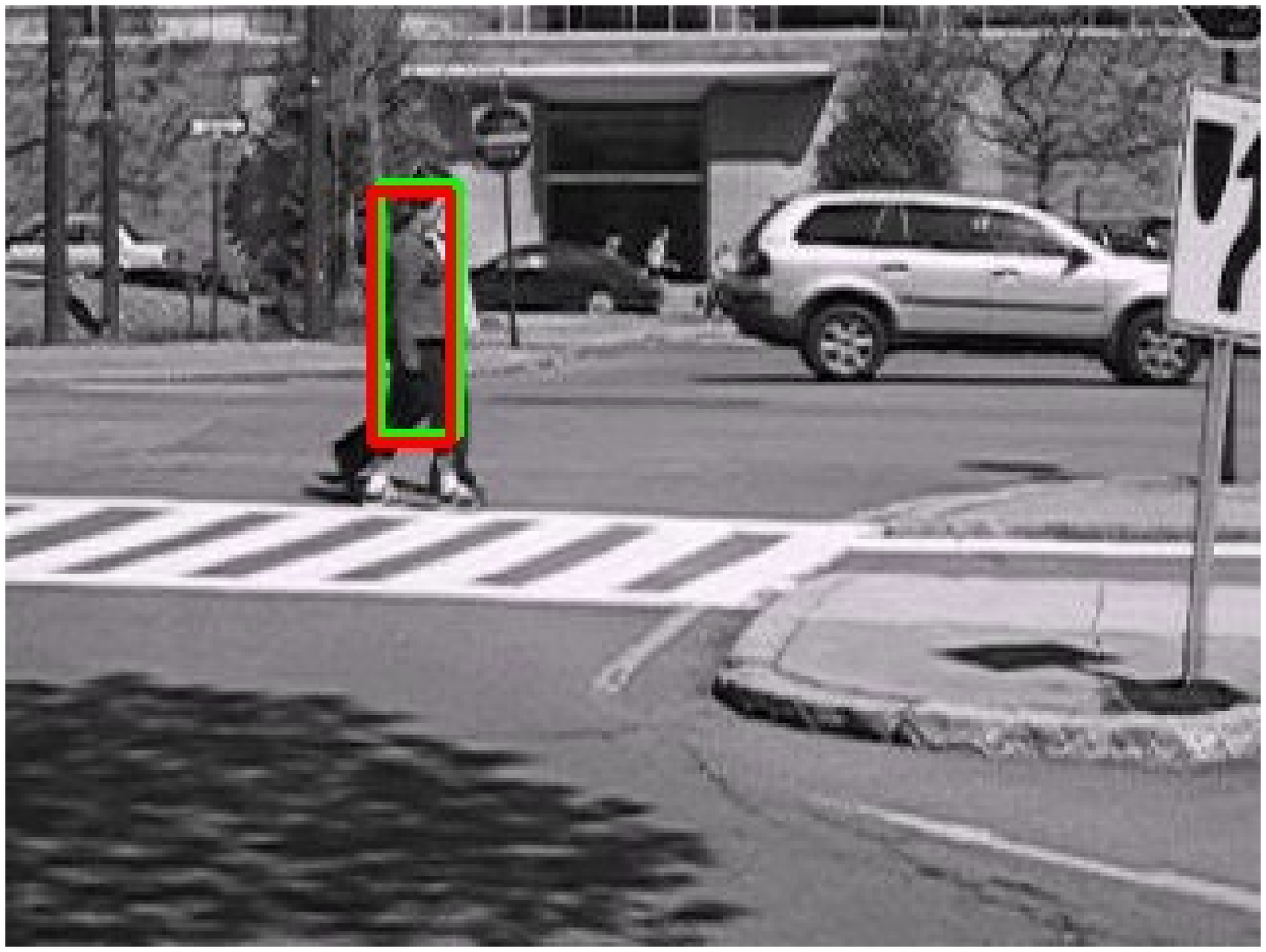}
}
\subfigure[Frame 106]{
\includegraphics[height=\imheight]{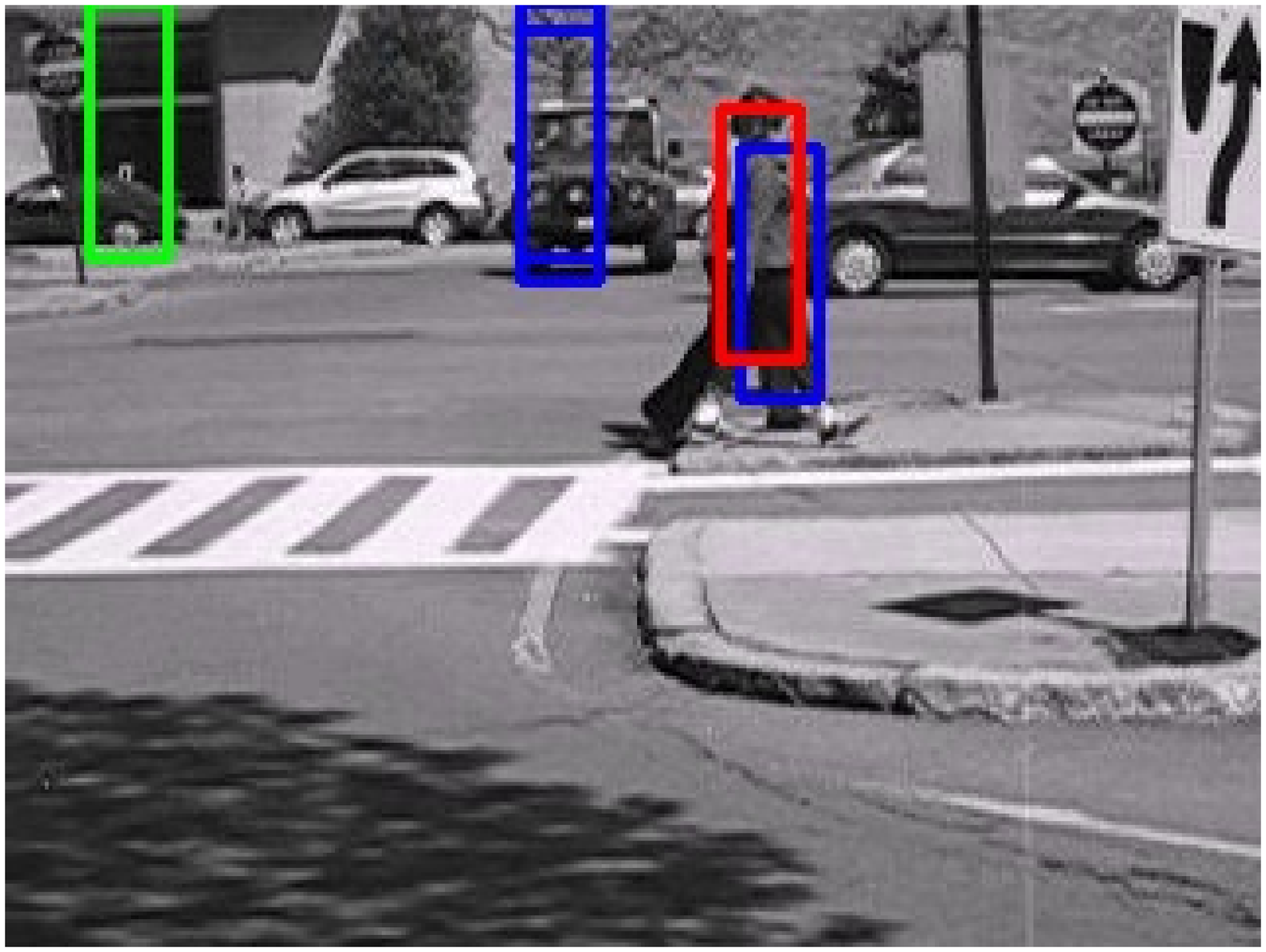}
}
\subfigure[Frame 140]{
\includegraphics[height=\imheight]{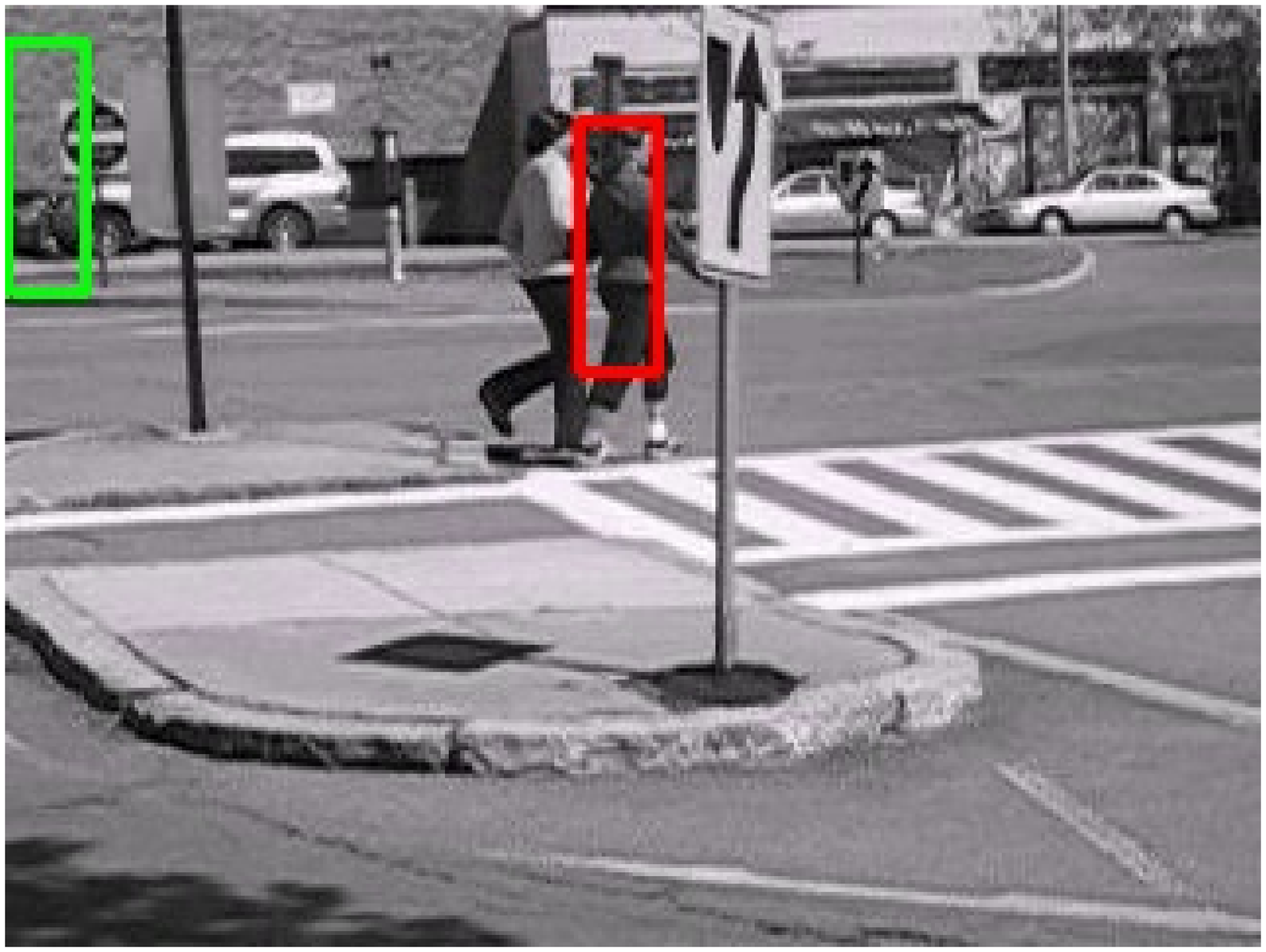}
}
\caption{Sequence {\em Crosswalk}: The frames 1, 17, 51, 75, 106and
140 are shown. The red boxes are tracked objects using DNBS, the green boxes are tracking results using NBS and the blue boxes
in some of the frames are sampled backgrounds for subspace update in DNBS.}
\label{fig:crosswalk}
\end{figure*}

\begin{figure*}
\def\imheight{2.8cm}
\centering
\subfigure[Frame 1]{
\includegraphics[height=\imheight]{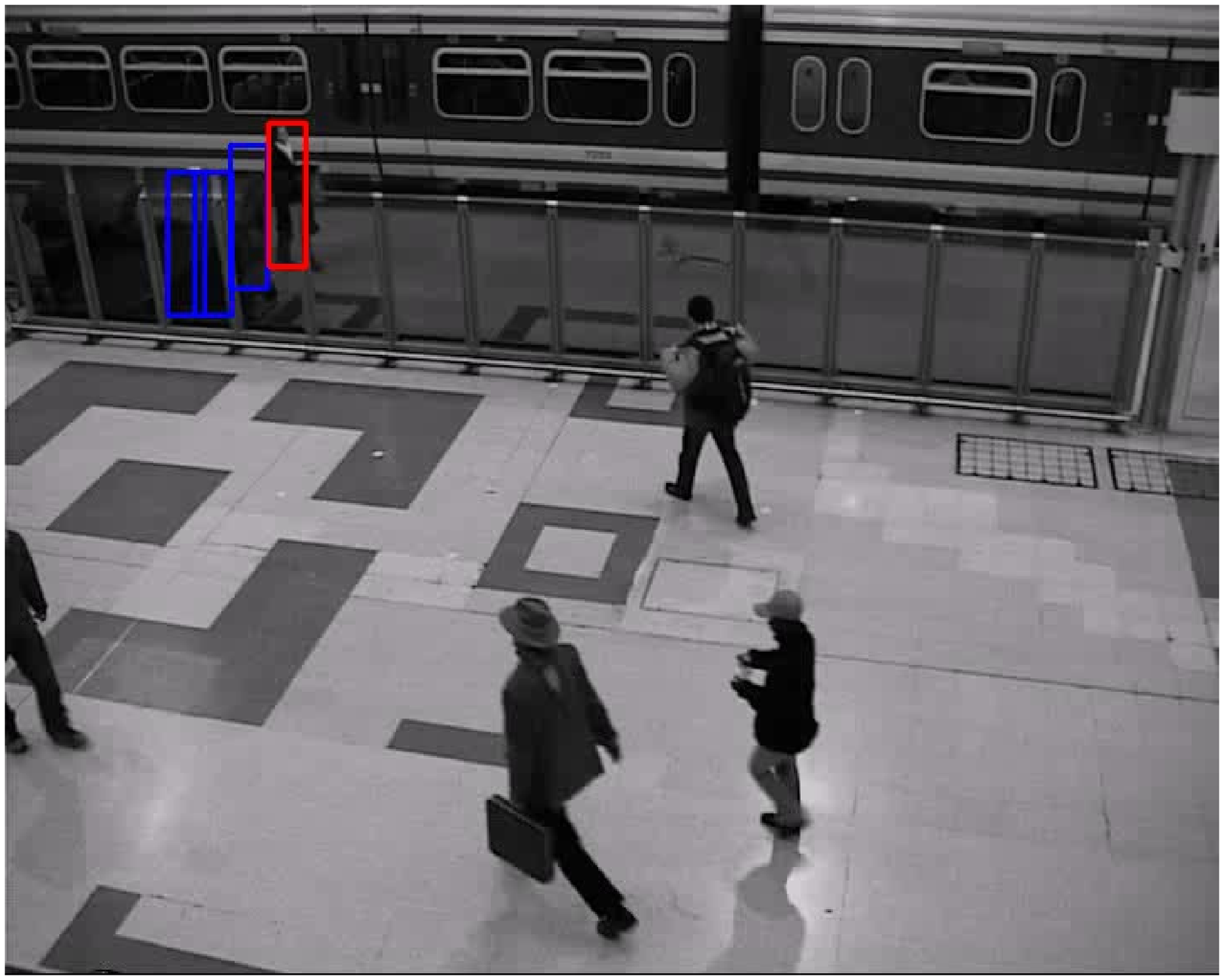}
}
\subfigure[Frame 41]{
\includegraphics[height=\imheight]{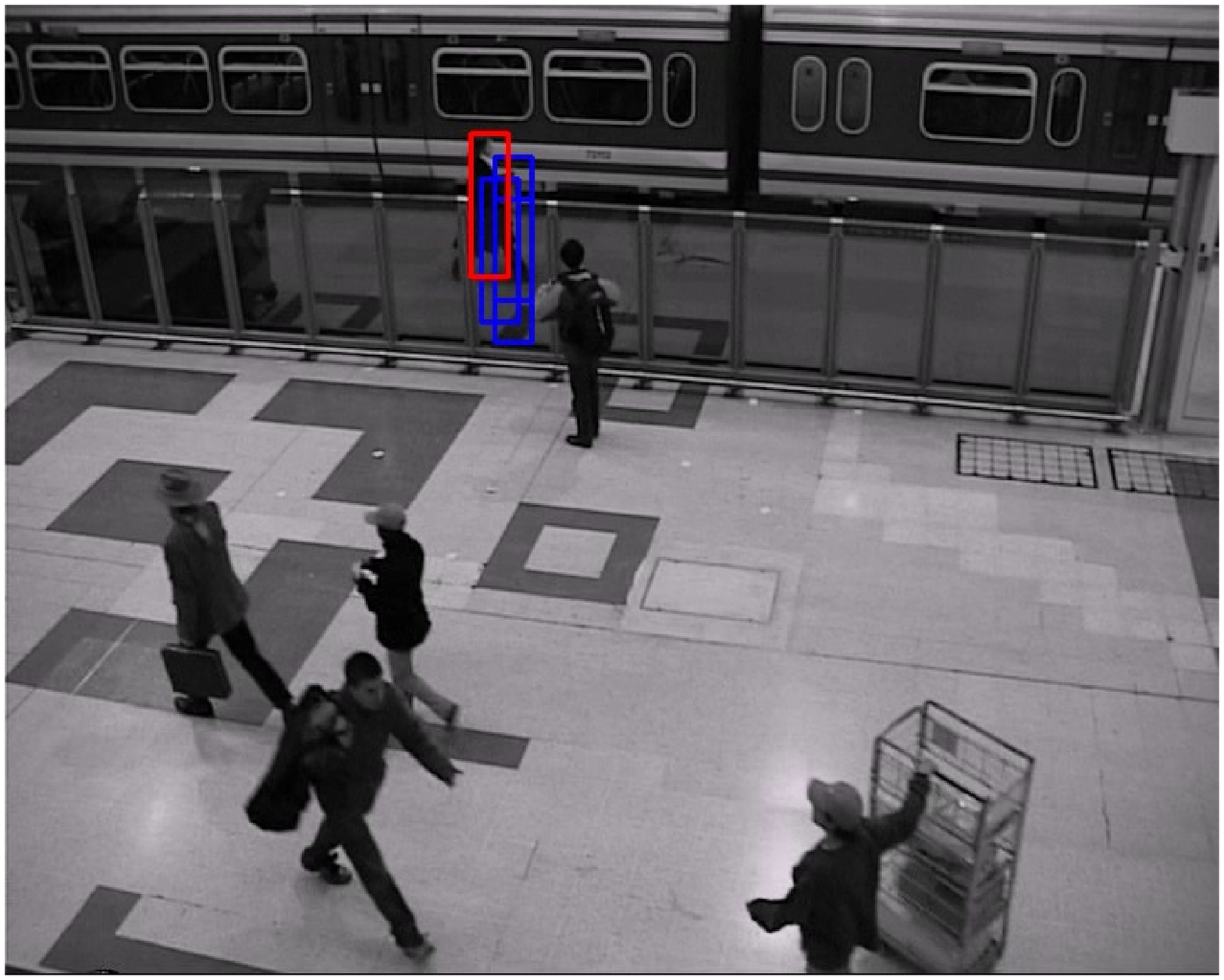}
}
\subfigure[Frame 76]{
\includegraphics[height=\imheight]{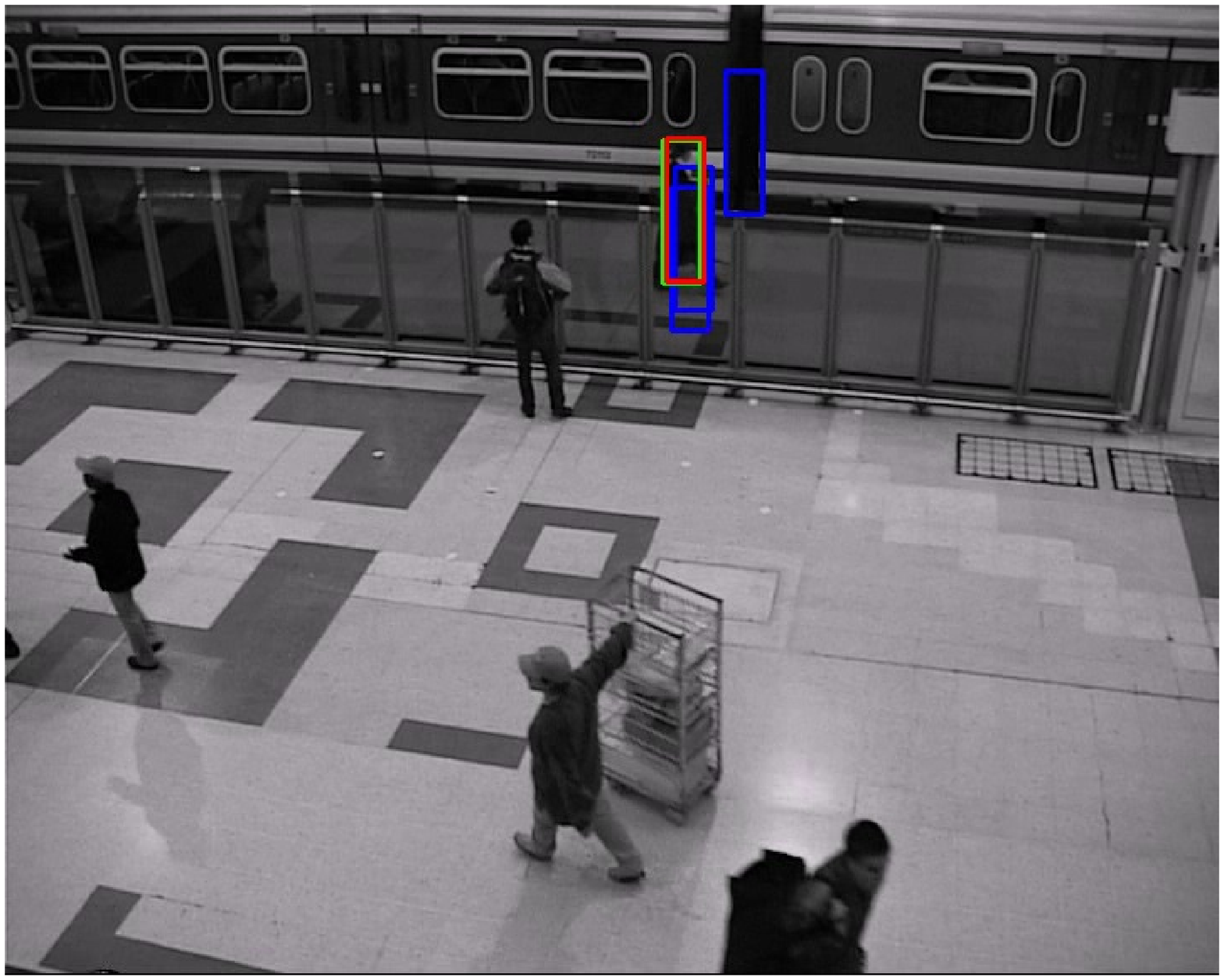}
}\\
\subfigure[Frame 91]{
\includegraphics[height=\imheight]{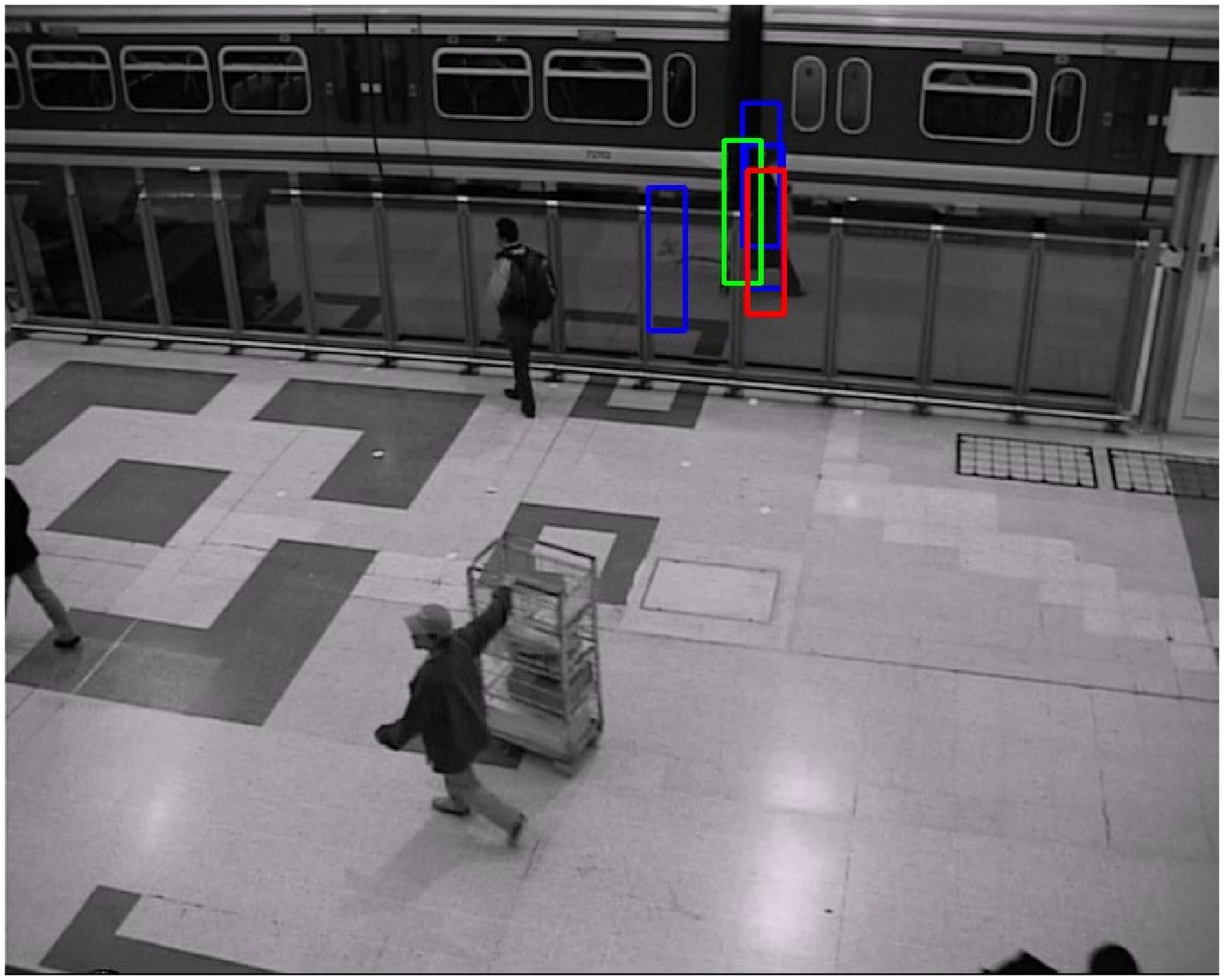}
}
\subfigure[Frame 107]{
\includegraphics[height=\imheight]{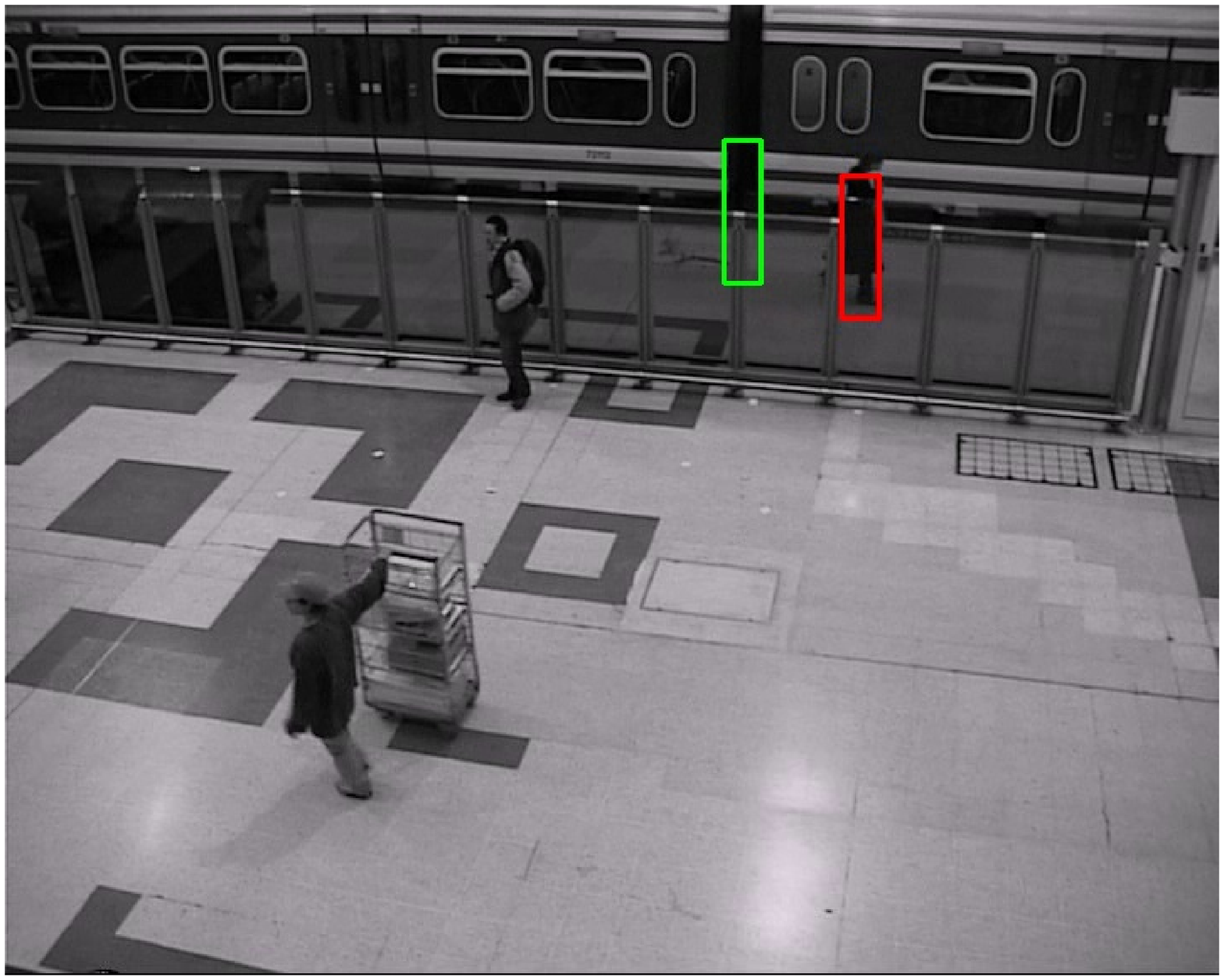}
}
\subfigure[Frame 153]{
\includegraphics[height=\imheight]{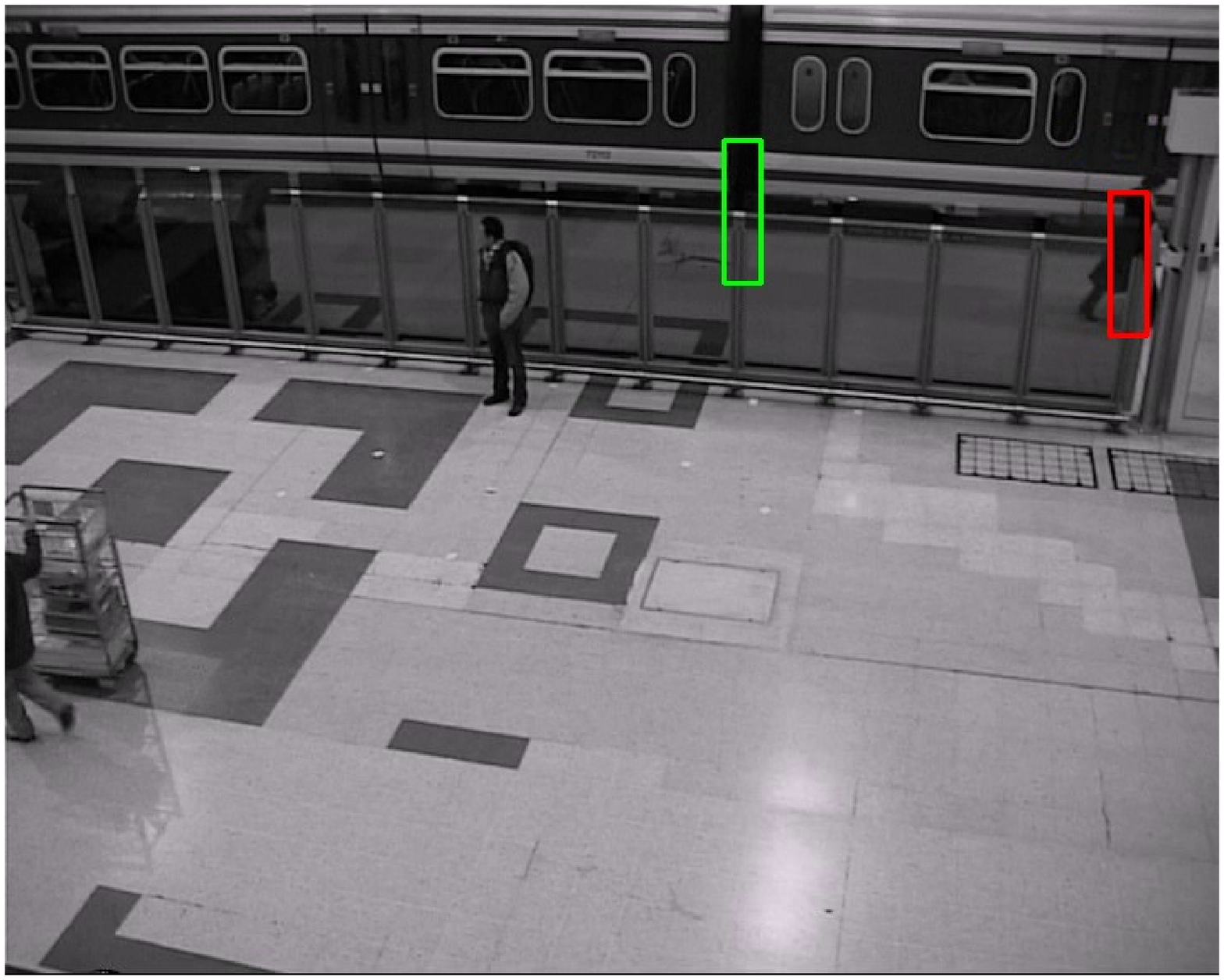}
}
\caption{Sequence {\em OccFemale}: The frames 1, 41, 76, 91, 107, and 153 are shown. The red boxes are tracked objects using DNBS, the green boxes are tracking results using NBS and the blue boxes
in some of the frames are sampled backgrounds for subspace update in DNBS.}
\label{fig:occfemale}
\end{figure*}

\begin{figure*}
\def\imheight{2.7cm}
\centering
\subfigure[Frame 1]{
\includegraphics[height=\imheight]{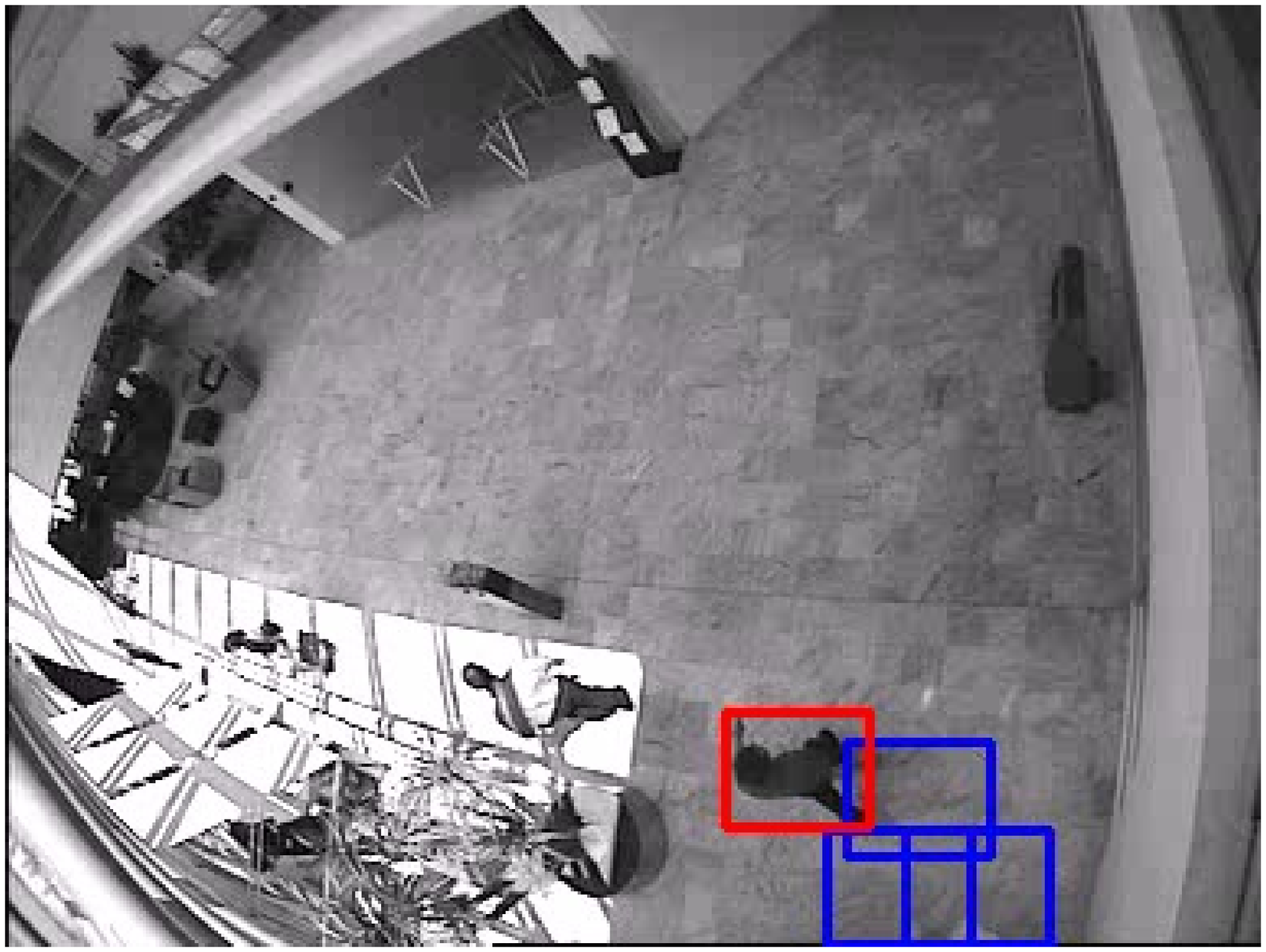}
}
\subfigure[Frame 22]{
\includegraphics[height=\imheight]{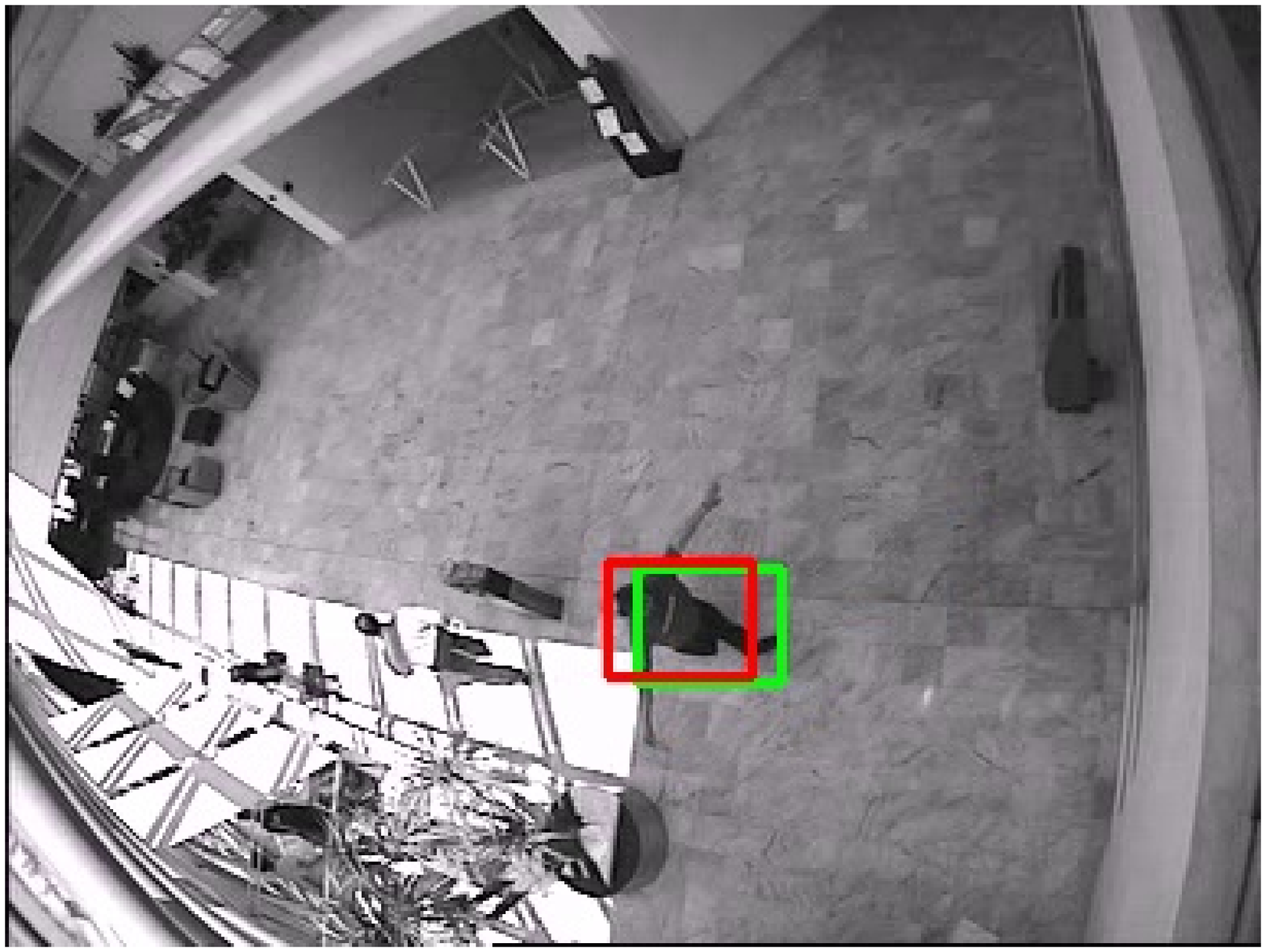}
}
\subfigure[Frame 52]{
\includegraphics[height=\imheight]{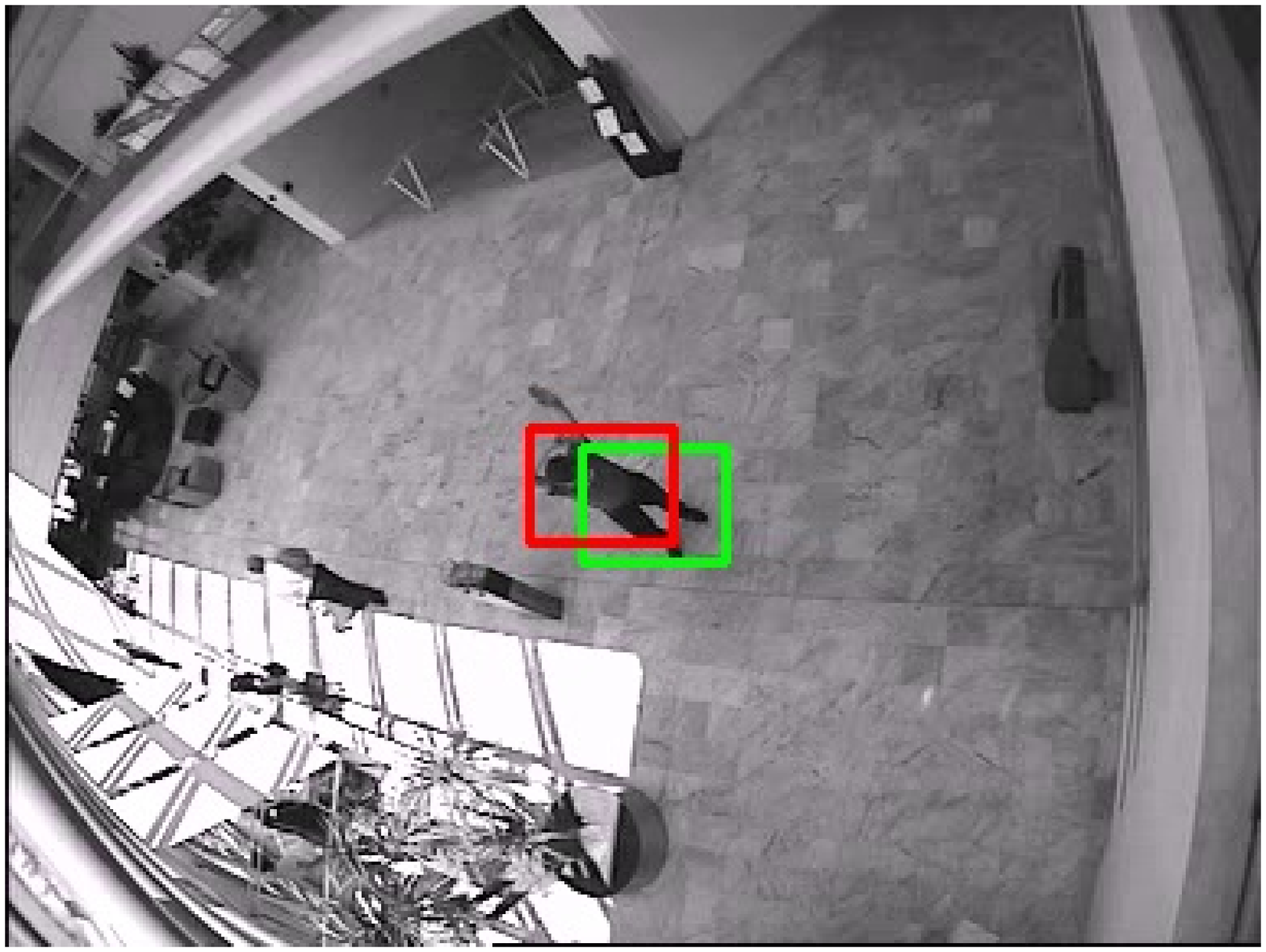}
}\\
\subfigure[Frame 92]{
\includegraphics[height=\imheight]{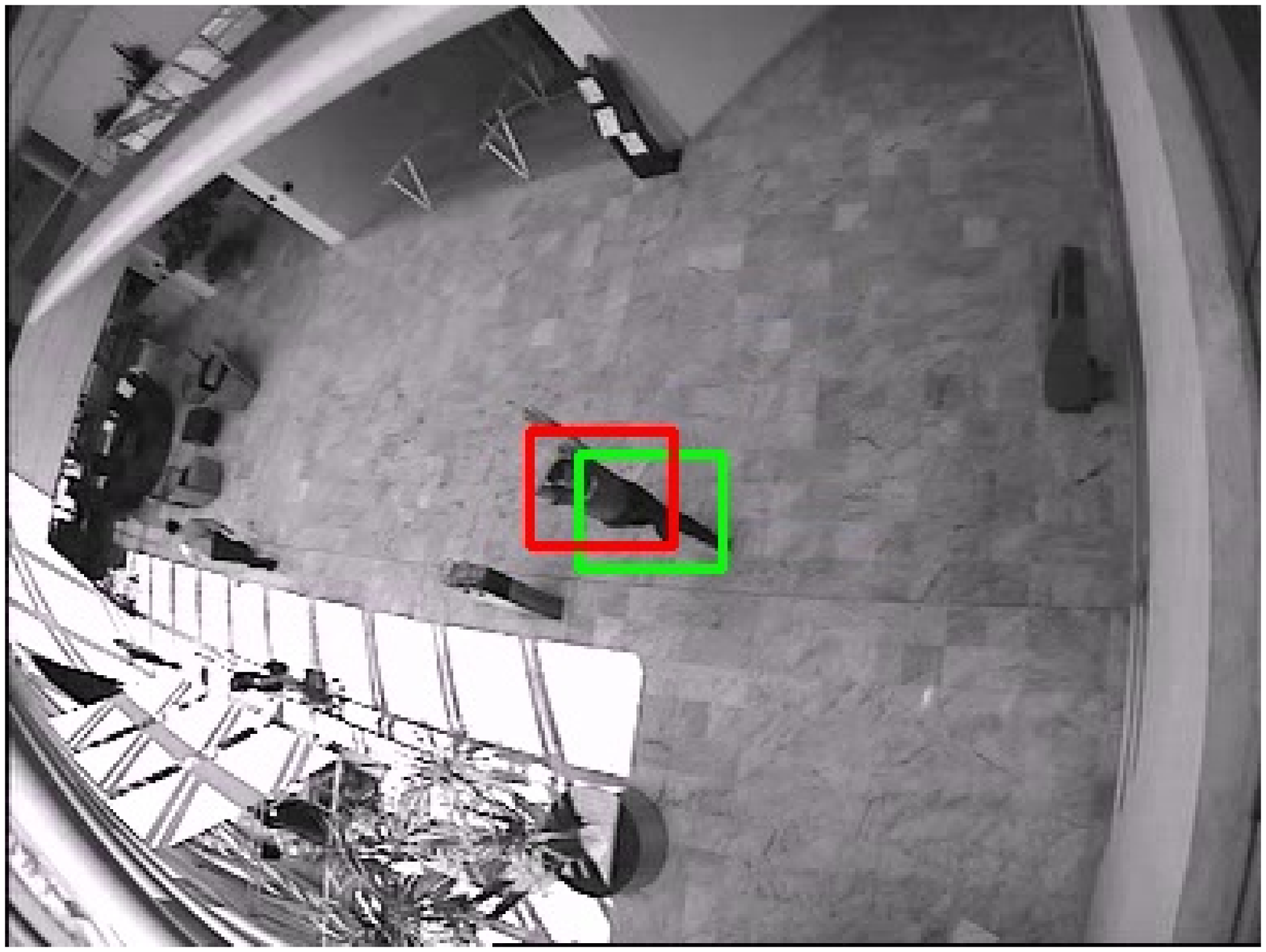}
}
\subfigure[Frame 116]{
\includegraphics[height=\imheight]{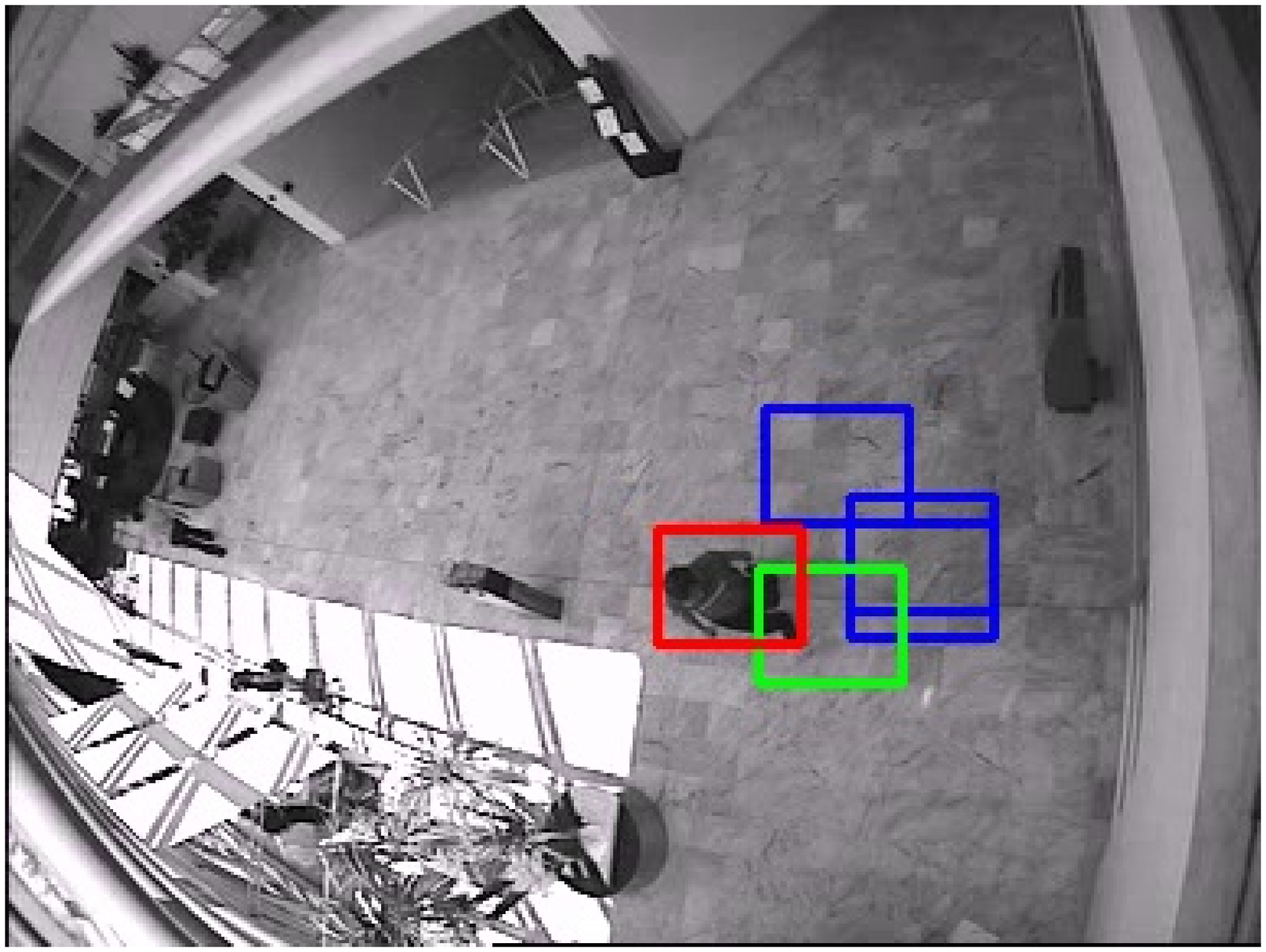}
}
\subfigure[Frame 142]{
\includegraphics[height=\imheight]{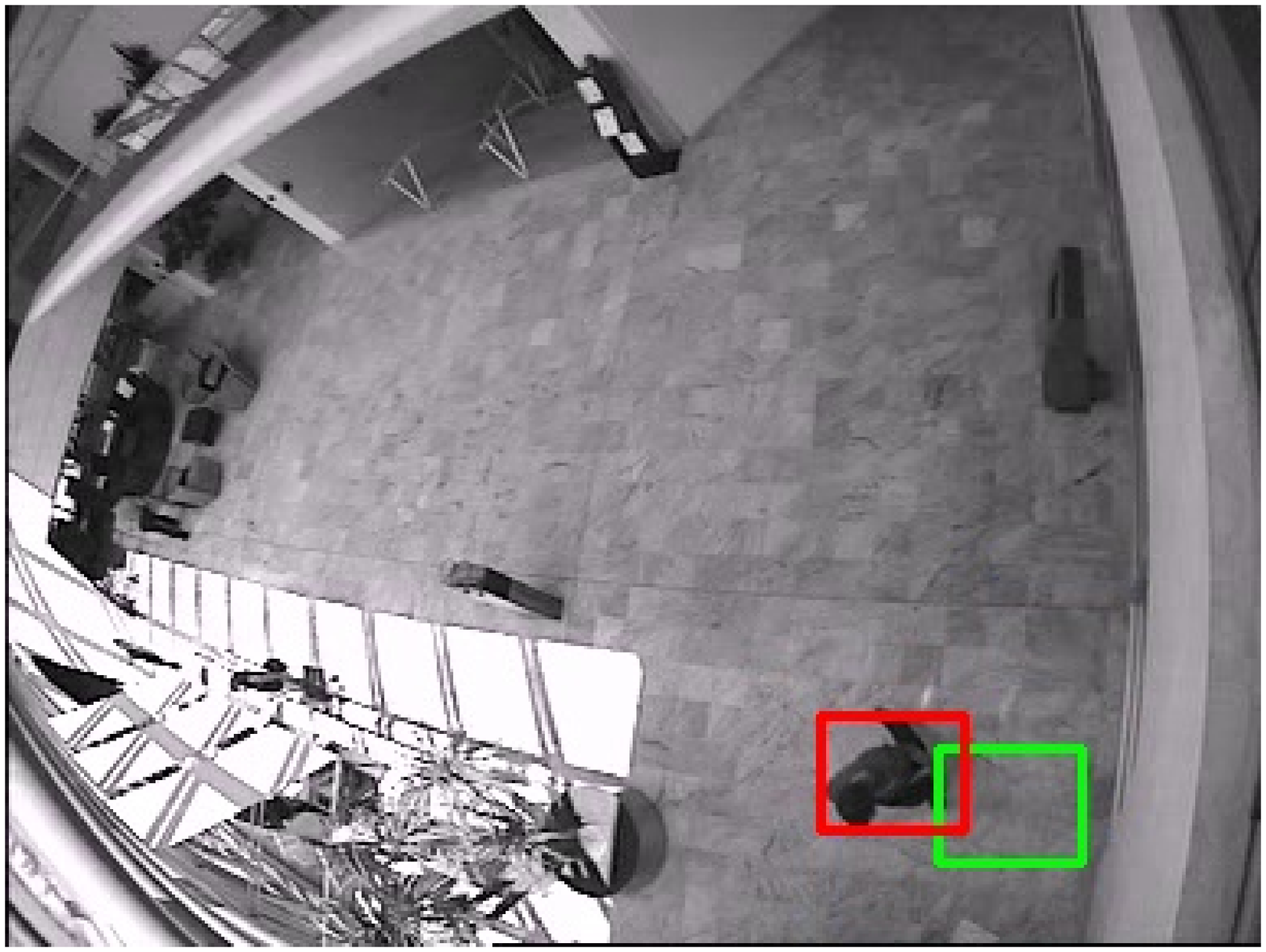}
}
\caption{Sequence {\em Browse} : The frames 1, 22, 52, 92, 116, 142
are shown. The red boxes are tracked objects using DNBS, the green boxes are tracking results using NBS and the blue boxes
in some of the frames are sampled backgrounds for subspace update in DNBS.} \label{fig:browse}
\end{figure*}

Fig. \ref{fig:timestats} shows the detailed computational cost with respect to the selection of $\mu$ for
tracking using hierarchical D-OOMP. The time statistics has three
components: (a) preprocessing the Haar-like dictionary and setting up relevant
parameters for tracking tasks (shown as the blue curve); (b) training, i.e., optimizing the DNBS
formulation to obtain the up-to-date DNBS subspace representation (shown as the green curve); (c)
tracking and localizing the foreground object in each frame (shown as red curve). The merit of
using Haar-like features is revealed in Fig. \ref{fig:timestats} that the tracking procedure
has an extremely low computation cost. Also, as the inner product upper limit $\mu$
changes we could find the optimal spot between 0.6 and 0.8 where the time
consumption for all of the three procedures is the minimum.

\subsection{Qualitative Results}
We apply our tracker to several challenging sequences to show its
effectiveness. Qualitative results are demonstrated on pedestrian videos
to show that our tracker can handle background clutter, heavy camera
motion, and object appearance variations. In the following figures,
red boxes indicate tracked object while blue ones are the
negative samples selected when the object DNBS is update at that
frame. The subspace is updated every 5 frames and if there is no
update of subspace. No blue boxes (background samples) will be
shown while no subspace update is performed.

We qualitatively compare the tracking results of the proposed DNBS approach with the NBS tracker to show the power of the additional discriminative term in Eq. \ref{dnbs:form}. To make the comparison fair, we fix the number of selected bases for both NBS and DNBS to be 30.

Sequence {\em Crowd} (Fig. \ref{fig:crowd}) is a video clip selected from PETS 2007 data set. In this sequence
the background is cluttered with many distracters. As can be
observed the object can still be well tracked. The frame is of size
$720\times 576$ and the object is initialized with a $26\times 136$
bounding box.

Sequence {\em Crosswalk} (Fig. \ref{fig:crosswalk}) has totally
140 frames, with two pedestrians walking together along a crowded
street with an extremely cluttered background. The tracking result
demonstrates the discriminative power of our algorithm. In this
sequence the hand-held camera is extremely unstable. The shaky
nature of the sequence makes it difficult to accurately
track the pedestrians. Despite this, our algorithm is able to track
the pedestrians throughout the entire 140 frames of the sequence.

Sequence {\em OccFemale} (Fig. \ref{fig:occfemale}) is a video clip selected
from the PETS 2006 data set. Each frame is of size $720\times576$ and
the object is initialized with a $22\times85$ bounding box at the beginning. It can be observed that the person being tracked is of low texture with very similar background and the person is also occluded by the fences periodically. In particular, the person's cloth is almost all black which makes it very similar to the black connector of the two compartments. When the person walks by the black connector at frame 90, the NBS tracker loses track (shown as a green box) while the DNBS tracker (shown as red box) can still keep track. This is because, at frame 76 this connector was selected as the background negative samples (the blue box) for model updating which makes the tracker aware of the distracting surroundings. The object can thus be tracked stably.

{
Sequence {\em Browse} (Fig. \ref{fig:browse}) is a video clip of
frames 24-201 extracted from {\em Browse1.avi} in CAVIAR people
(ECCV-PETS 2004) dataset\footnote{CAVIAR Dataset, EC Funded CAVIAR project: \url{http://homepages.inf.ed.ac.uk/rbf/CAVIAR/}}. This sequence is
recorded by a distorted camera. Each frame is $384\times 288$ pixels
and the object is bounded by a $44\times 35$ box. With significant
distortion, the object can still be tracked. }

In addition, we validated our tracker on other sequences from public video datasets such as 
Sequence \textit{Courtyard}, Sequence \textit{Ferry} which is extracted from PETS 2005 Zodiac Dataset\footnote{PETS 2005 Zodiac: \url{http://www.vast.uccs.edu/~tboult/PETS05/}}, Sequence \textit{CrowdFemale} extracted from PETS 2007\footnote{PETS 2007: \url{http://www.cvg.reading.ac.uk/PETS2007/data.html}}, and sequences \textit{boy}, \textit{car4}, \textit{couple}, \textit{crossing}, \textit{david}, \textit{david2}, \textit{fish}, \textit{girl}, \textit{matrix}, \textit{mhyang}, \textit{soccer}, \textit{suv}, \textit{trellis} which are used in previous literatures \cite{WuLimYang13}.  {Qualitative video results for all of these 21 sequences are available at our demo webpage\footnote{DNBS webpage: \url{http://www.cs.umd.edu/users/angli/dnbs}}.

\begin{table*}[!htb]
\caption{Performance evaluation using the fraction of successfully tracked frames. A frame is successfully tracked if and only if the overlap ratio, i.e. intersection over union (IOU), is higher than 0.35, with respect to the ground truth bounding box. The timing of each method (frames per second) is computed with respect to a $44\times35$ object template. For each sequence, the best success fraction is highlighted in red color. Our tracker ranks 1st both in the averaged success rate and in the total number of winning sequences with a moderate real time speed compared to all other trackers.}
\label{tab1}
\centering
\begin{tabular}{|c|c|c|c|c|c|c|c|c|c||c|c|}
\hline
\multirow{2}{*}{\bf \#} & \multirow{2}{*}{\bf Sequence}            & \multirow{2}{*}{\bf CSK}  & \multirow{2}{*}{\bf CT}   & \multirow{2}{*}{\bf DFT}  & \multirow{2}{*}{\bf IVT}  & \multirow{2}{*}{\bf L1APG} & \multirow{2}{*}{\bf ORIA} & \multirow{2}{*}{\bf Struck} & \multirow{2}{*}{\bf TLD}  & \multicolumn{2}{|c|}{\bf Proposed}  \\ \cline{11-12}
 &   & & & & & & & & & \bf NBS & \bf DNBS \\\hline\hline
1 & blackman & 0.55 & 0.01 & \win 1.00 & \win 1.00 & \win 1.00 & 0.13 & 0.99 & 0.32 & \win 1.00 & \win 1.00\\ \hline
2 & boy & 0.84 & 0.64 & 0.48 & 0.33 & 0.93 & 0.18 & \win 1.00 & \win 1.00 & 0.43 & 0.44\\ \hline
3 & browse & 0.85 & 0.39 & 0.88 & 0.15 & \win 0.90 & 0.07 & 0.70 & 0.09 & 0.87 & 0.87\\ \hline
4 & car4 & 0.67 & 0.35 & 0.26 & \win 1.00 & 0.32 & 0.24 & 0.73 & 0.27 & 0.28 & 0.30\\ \hline
5 & couple & 0.09 & 0.32 & 0.09 & 0.09 & 0.61 & 0.05 & 0.71 & 0.25 & 0.54 & \win 0.99\\ \hline
6 & courtyard & \win 1.00 & \win 1.00 & \win 1.00 & \win 1.00 & \win 1.00 & 0.34 & \win 1.00 & \win 1.00 & 0.96 & \win 1.00\\ \hline
7 & crossing & 0.88 & 0.99 & 0.68 & 0.40 & 0.25 & 0.21 & \win 1.00 & 0.52 & 0.37 & 0.71\\ \hline
8 & crosswalk & 0.06 & 0.09 & 0.09 & 0.09 & 0.10 & 0.01 & 0.66 & 0.12 & 0.55 & \win 0.99\\ \hline
9 & crowd & \win 1.00 & 0.01 & \win 1.00 & \win 1.00 & \win 1.00 & 0.47 & \win 1.00 & 0.57 & \win 1.00 & \win 1.00\\ \hline
10 & crowdfemale & 0.16 & 0.95 & \win 1.00 & 0.99 & \win 1.00 & 0.42 & 0.97 & 0.74 & 0.49 & \win 1.00\\ \hline
11 & david & 0.48 & 0.90 & 0.35 & \win 0.96 & 0.81 & 0.47 & 0.31 & 0.63 & 0.84 & 0.88\\ \hline
12 & david2 & \win 1.00 & 0.00 & 0.60 & \win 1.00 & \win 1.00 & 0.70 & \win 1.00 & \win 1.00 & \win 1.00 & \win 1.00\\ \hline
13 & ferry & 0.28 & 0.27 & 0.29 & 0.27 & 0.27 & 0.05 & 0.30 & 0.81 & \win 0.99 & \win 0.99\\ \hline
14 & fish & 0.04 & 0.97 & 0.86 & \win 1.00 & 0.26 & 0.70 & \win 1.00 & 0.68 & \win 1.00 & \win 1.00\\ \hline
15 & girl & 0.48 & 0.35 & 0.29 & 0.21 & 0.99 & 0.55 & \win 1.00 & 0.87 & 0.78 & 0.70\\ \hline
16 & matrix & 0.01 & 0.10 & 0.06 & 0.02 & 0.11 & \win 0.23 & 0.19 & 0.12 & 0.02 & 0.02\\ \hline
17 & mhyang & \win 1.00 & 0.86 & 0.94 & \win 1.00 & \win 1.00 & \win 1.00 & \win 1.00 & \win 1.00 & \win 1.00 & \win 1.00\\ \hline
18 & occfemale & 0.56 & \win 1.00 & 0.99 & 0.99 & 0.83 & 0.06 & 0.85 & 0.61 & 0.57 & 0.99\\ \hline
19 & soccer & 0.16 & \win 0.34 & 0.23 & 0.16 & 0.21 & 0.17 & 0.16 & 0.15 & 0.24 & 0.24\\ \hline
20 & suv & 0.59 & 0.24 & 0.06 & 0.46 & 0.55 & 0.59 & 0.73 & \win 0.98 & 0.53 & 0.57\\ \hline
21 & trellis & \win 0.85 & 0.40 & 0.53 & 0.37 & 0.31 & 0.61 & 0.63 & 0.43 & 0.56 & \win 0.85\\ \hline\hline
\multicolumn{2}{|c|}{averaged success rate} & 0.55 & 0.48 & 0.56 & 0.59 & 0.64 & 0.35 & 0.76 & 0.58 & 0.67 & \win 0.79\\ \hline
\multicolumn{2}{|c|}{\# winning sequences} & 5 & 3 & 4 & 8 & 7 & 2 & 8 & 5 & 6 & \win 11\\ \hline
\multicolumn{2}{|c|}{frames per second}     & \win 342  & 76   & 10   & 26   & 1     & 14   & 7      & 24   & 22 & 17        \\ \hline
\end{tabular}
\end{table*}

\def\picwidth{0.12\textwidth}
\subsection{Quantitative Evaluation}

In the quantitative evaluation, we compare our tracker with 8 state-of-the-art trackers, which are CSK \cite{henriques2012circulant}, CT \cite{ZhangCT}, DFT \cite{DFT}, IVT \cite{ivt}, L1APG \cite{L1APG}, ORIA \cite{ORIA}, Struck \cite{Struck} and TLD \cite{TLD}, using 21 public video sequences.

In the first place, we give a detailed comparison in Table \ref{tab1} where each of the trackers is evaluated on the same set of 21 video sequences. Each cell in the the table shows the percentage of successfully tracked frames with respect to the corresponding tracker-sequence pair. A frame is successfully tracked if and only if the overlap ratio (intersection area over union area) between tracked object and ground truth is higher than 0.35, i.e. more than half portion of the object is overlapped with the groundtruth bounding box. For each sequence, the highest success fractions are highlighted in red color. The last three rows show respectively the averaged success rate, the number of winning sequences, and the frames per second. The average success rate is computed by averaging the success rate for each of the sequences. According to the result, the proposed DNBS tracker performs the best (0.79 in average success rate) and Struck ranks 2nd with a very close success rate 0.76. The number of winning sequences is computed by counting the total number of sequences where each of the methods wins. The proposed DNBS approach wins in total 11 sequences (Blackman, Couple, Courtyard, Crosswalk, Crowd, Crowdfemale, Ferry, Fish, Mhyang, Occfemale and Trellis) which outperforms all the other trackers. Struck and IVT tied for the 2nd place with 8 winings. The number of frames tracked per second (FPS) is shown in the last row which is calculated using an object template of size $44\times35$. Our DNBS tracker runs in real-time at 17 frames per second which is faster than DFT, L1APG, ORIA and Struck.

For a more comprehensive evaluation, we employ the quantitative evaluation protocols proposed by \cite{WuLimYang13}. There are three criteria used in their evaluation benchmark: (a) one-pass evaluation (OPE) tests each tracker from the beginning of the sequence to the end; (b) spatial robustness evaluation (SRE) initializes the tracker 12 times on each sequence by spatially pertubing the initial bounding box and averages the performance of different initializations over all trials; and (c) temporal robustness evaluation (TRE) segments each sequence into 20 segments and tests the tracker on each segment independently and averages their performances over all trials. Besides, two error functions are employed: the centeroid distance from the tracked object location to the ground-truth location and the bounding box overlap (Intersection-Over-Union) ratio. Such comprehensive criteria provide a better evaluation of the robustness of trackers. The resulting precision curves using overlap rate are shown in Fig. \ref{fig:quanall-overlap} and curves using centroid distance error are shown in Fig. \ref{fig:quanall}. According to the curves, the performance of our tracker is the best in OPE evaluation with overlap ratio, and ranks 2nd with any of the other 5 evaluation criteria.

\begin{figure*}
\centering
\subfigure[]{
\includegraphics[width=0.32\textwidth]{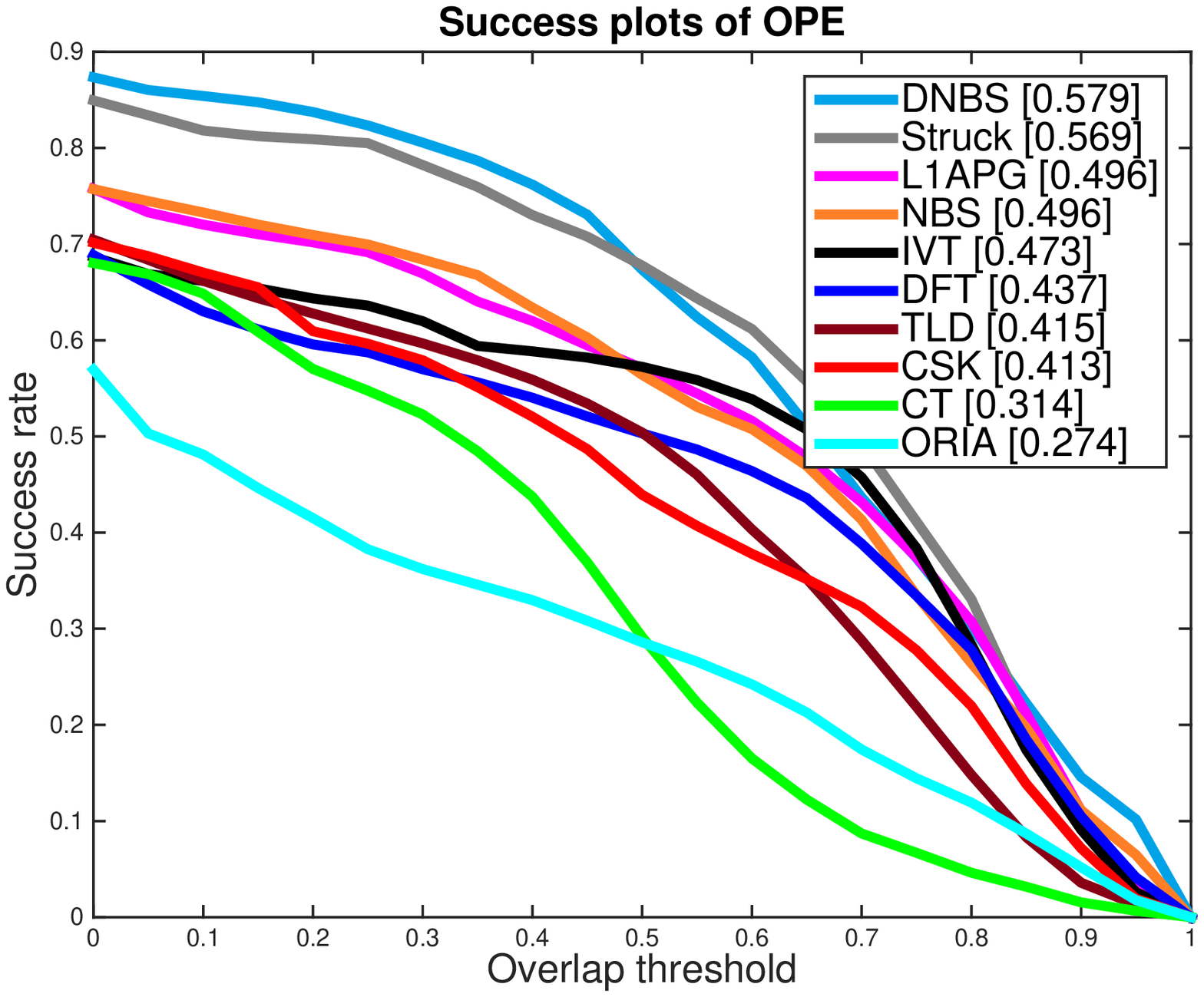}
}\subfigure[]{
\includegraphics[width=0.32\textwidth]{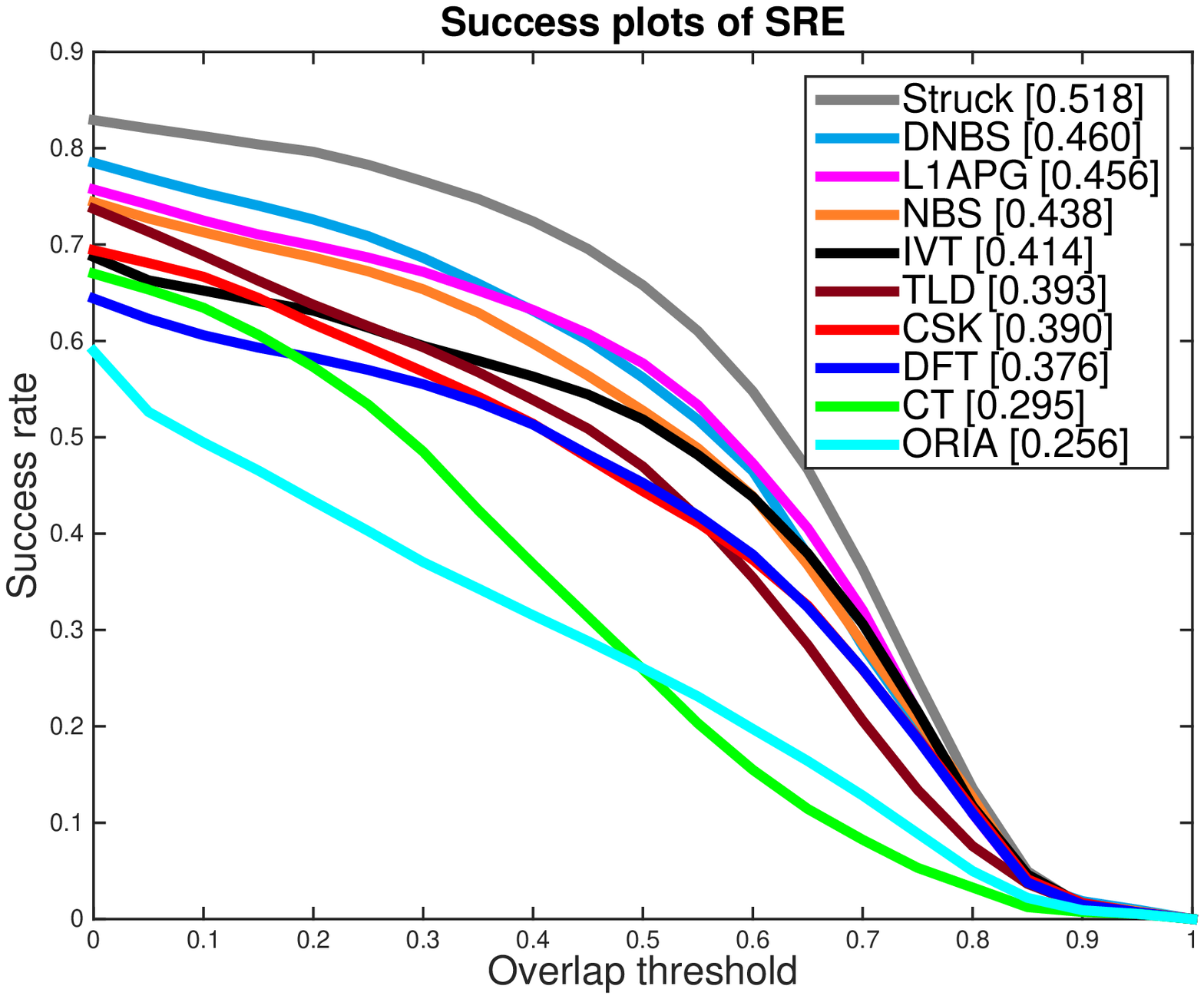}
}\subfigure[]{
\includegraphics[width=0.32\textwidth]{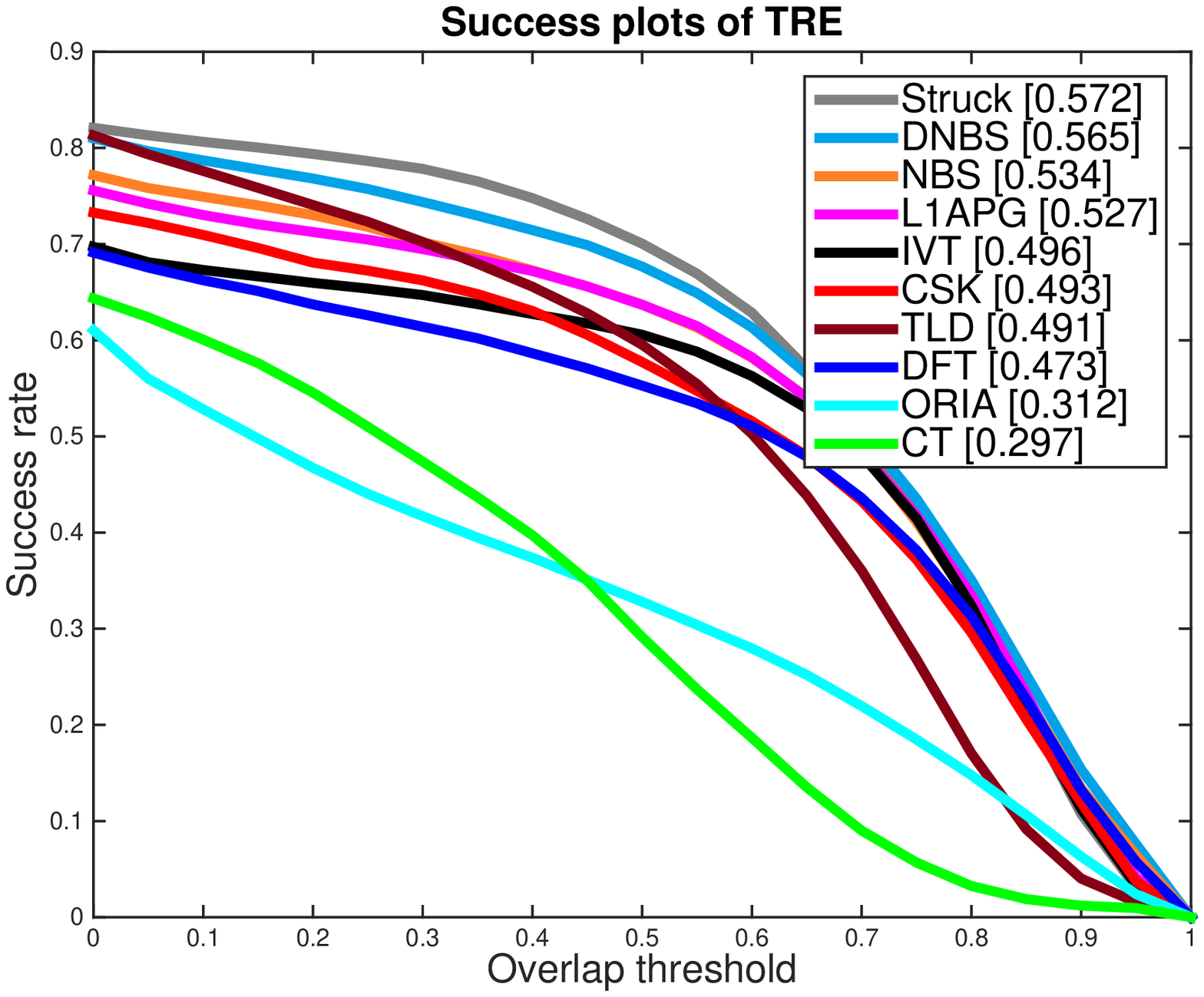}
}
\caption{Quantitative results on the success rate with respect to overlap ratio over 21 sequences: (a) One-pass evaluation (b) Spatial-robustness evaluation (c) Temporal-robustness evaluation}\label{fig:quanall-overlap}
\end{figure*}

\begin{figure*}
\centering
\subfigure[]{
\includegraphics[width=0.32\textwidth]{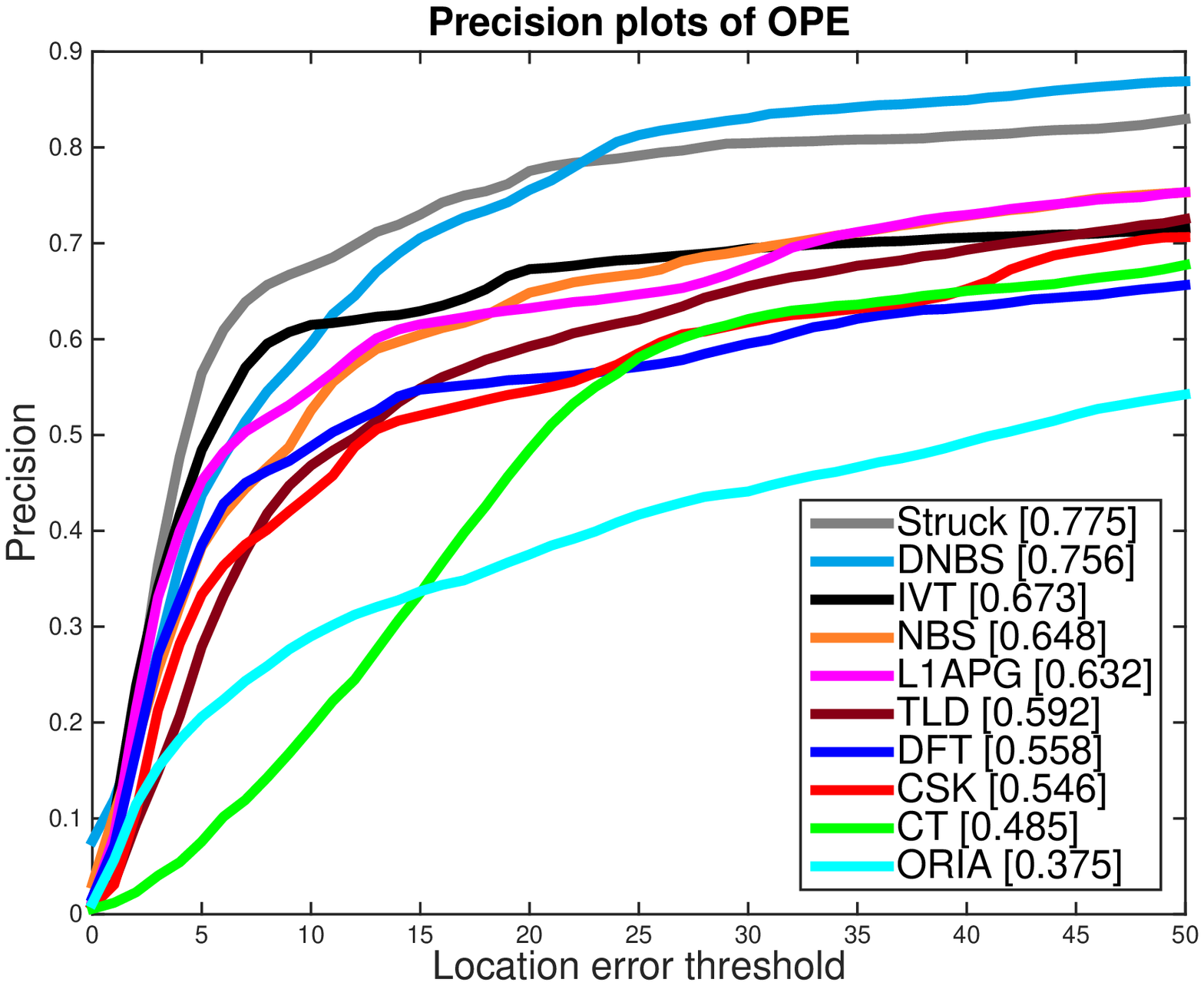}
}\subfigure[]{
\includegraphics[width=0.32\textwidth]{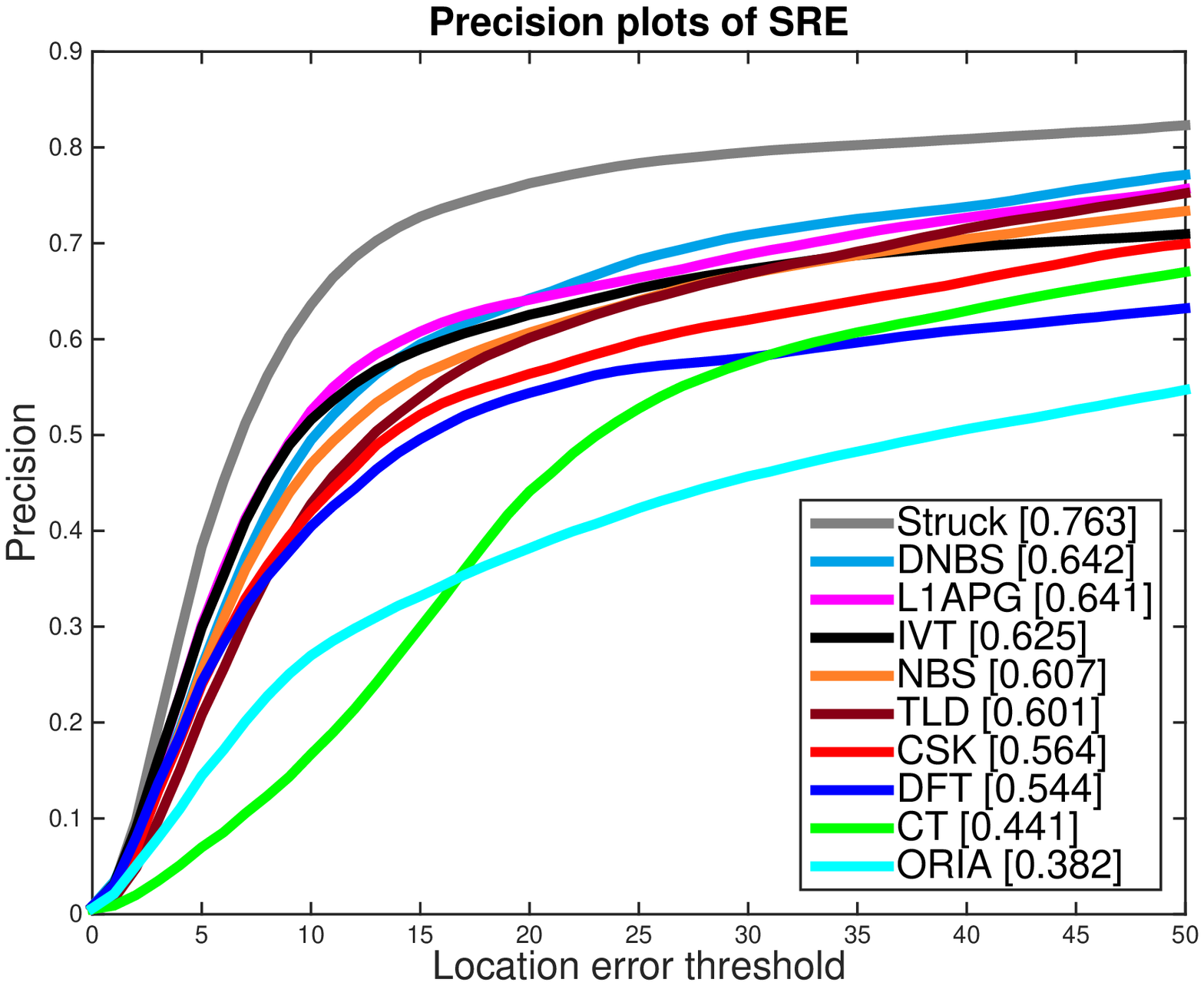}
}\subfigure[]{
\includegraphics[width=0.32\textwidth]{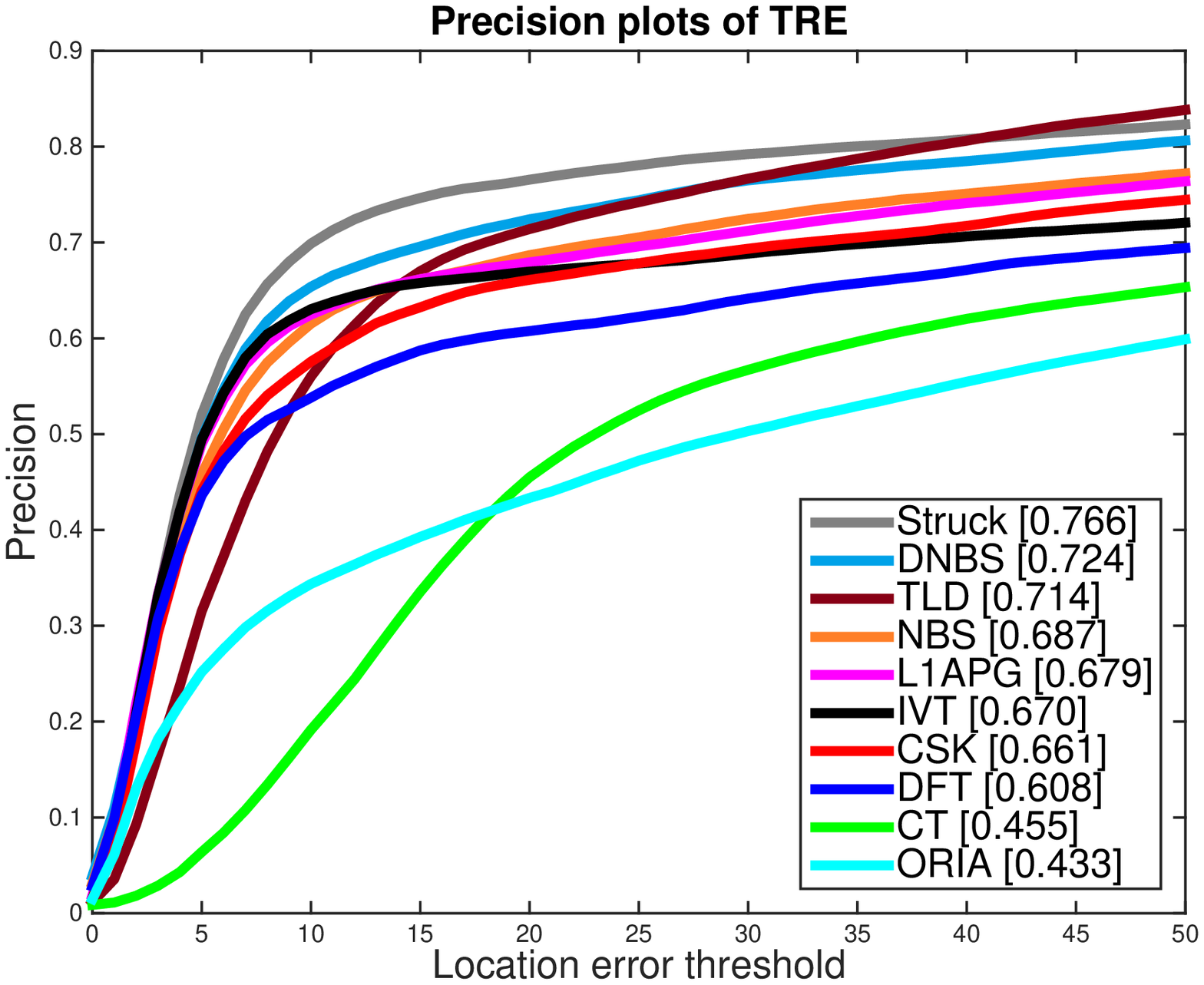}
}
\caption{Quantitative results on the object center distance error evaluations over 21 sequences: (a) One-pass evaluation (b) Spatial-robustness evaluation (c) Temporal-robustness evaluation}\label{fig:quanall}
\end{figure*}

According to quantitative results, it is hard to say that there is a tracker performs best in both computation cost and accuracy. In Table \ref{tab1}, our DNBS tracker outperforms other trackers in 11 of these sequences and its speed is faster than half of the trackers. The second most accurate tracker is Struck. However, in the comprehensive evaluation (Fig.~\ref{fig:quanall}), Struck performs the best for 5 of the criteria for which our DNBS tracker is the second. We have to admit that Struck provides more robustness in the application of object tracking. However, its speed is twice slower than our DNBS tracker. Since our method is based on template matching, we observed that it is hard for our tracker to compete with Struck in every scenario, which is based on structural learning. When compared to methods using similar techniques, our DNBS tracker outperforms all those subspace representation based approaches used in our evaluation such as IVT, CT, and L1APG. Since our approach does not rely on partile filtering, it provides robustness and efficiency in videos with heavy camera motion. The drawback of our approach is that it does not very well handle scale changes and non-rigid intra-object motions which is due to the nature of template matching. After all, our tracker is based on much simpler principles and algorithms which produces a relatively balanced performance in both accuracy and computation.

\section{Conclusion}\label{sec:conclusion}
We have proposed the Discriminative Nonorthogonal Binary Subspace, a simple yet informative
object representation that can be solved using a variant of OOMP.
The proposed DNBS representation incorporates the discriminate image
information to distinguish the foreground and background, making it
suitable for object tracking. We used SSD matching built
upon the DNBS to efficiently locate object in videos. The optimization of DNBS is efficient as we proposed a suite of algorithms to accelerate the training process.
Our experiments on challenging video sequences show that the
DNBS-based tracker can stably track the dynamic
objects. In the future, we intend to explore the applications of DNBS
on other computer vision and multimedia tasks such as image copy detection and face verification.

\bibliographystyle{IEEEtran}      
\bibliography{dnbs}   

\ifCLASSOPTIONcompsoc
  \section*{Acknowledgments}
\else
  \section*{Acknowledgment}
\fi

The authors would like to thank Dr. Terry Boult for sharing the Zodiac video dataset.

\ifCLASSOPTIONcaptionsoff
  \newpage
\fi

\appendix[{Proof of Lemma 1}\label{app:theo1}]
\begin{proof}
According to the property of inner product, we have
\begin{equation}
\langle\mathbf{x},R_\mathbf{\Phi}(\mathbf{y})\rangle=\langle R_\mathbf{\Phi}(\mathbf{x}),R_\mathbf{\Phi}(\mathbf{y})\rangle~.
\end{equation}
Since $\gamma_i^{(k)}=\mathbf{\psi}_i-R_{\mathbf{\Phi}_{k-1}}(\mathbf{\psi}_i)$ and $\varepsilon_{k-1}(\mathbf{x}) =
\mathbf{x}-R_{\mathbf{\Phi}_{k-1}}(\mathbf{x})$, hence
\begin{align}
\langle\gamma_i^{(k)},\varepsilon_{k-1}(\mathbf{x})\rangle&=\langle\mathbf{\psi}_i-R_{\mathbf{\Phi}_{k-1}}(\mathbf{\psi}_i), \mathbf{x}-R_{\mathbf{\Phi}_{k-1}}(\mathbf{x}) \rangle\nonumber\\
&=\langle\mathbf{\psi}_i,\mathbf{x}-R_{\mathbf{\Phi}_{k-1}}(\mathbf{x})\rangle\nonumber\\
&\quad-\langle R_{\mathbf{\Phi}_{k-1}}(\mathbf{\psi}_i), \mathbf{x}-R_{\mathbf{\Phi}_{k-1}}(\mathbf{x}) \rangle\nonumber\\
&=\langle\mathbf{\psi}_i,\mathbf{x}-R_{\mathbf{\Phi}_{k-1}}(\mathbf{x})\rangle~.
\end{align}
\end{proof}

\appendix[Proof of Lemma 2]
\begin{proof}
Only the orthogonal component of the newly added basis with respect to the old subspace is able to contribute to the update of the image reconstruction, therefore
\begin{equation}\label{lemma2:eq}
R_{\mathbf{\Phi}_{k}}(\mathbf{x})=R_{\mathbf{\Phi}_{k-1}}(\mathbf{x})+\frac{\varphi_k\langle\varphi_k,\mathbf{x}\rangle}{\parallel\varphi_k\parallel^2}
~,
\end{equation}
where $\varphi_k=\phi_k-R_{\mathbf{\Phi}_{k-1}}(\phi_k)$ denotes the
component of $\phi_k$ that is orthogonal to the subspace spanned by
$\mathbf{\Phi}_{k-1}$, we therefore have the recursive definition of the $l^2$-norm of $\gamma_i^{(k)}$:
 \begin{align}
\parallel\gamma_i^{(k)}\parallel^2&=\parallel\mathbf{\psi}_i-R_{\mathbf{\Phi}_{k-2}}(\mathbf{\psi}_i)-\frac{\varphi_{k-1}\langle\varphi_{k-1},\mathbf{\psi}_i\rangle}{\parallel\varphi_{k-1}\parallel^2}\parallel^2\nonumber\\
&=\parallel\gamma_i^{(k-1)}\parallel^2-\frac{|\langle\varphi_{k-1},\mathbf{\psi}_i\rangle|^2}{\parallel\varphi_{k-1}\parallel^2}
~.
\end{align}
\end{proof}

\appendix[{Proof of Proposition 1}\label{theo1:proof}]
\begin{proof}
According to Eq. \ref{lemma2:eq}, the inner product between residues of Haar-like feature and image vector is
\begin{equation}
\begin{split}
\langle\gamma_i^{(k)}, & \varepsilon_{k-1}(\mathbf{x})\rangle=\\
&\langle\gamma_i^{(k-1)},\varepsilon_{k-2}(\mathbf{x})\rangle-\frac{\langle\varphi_{k-1},\psi_i\rangle\cdot\langle\varphi_{k-1},\mathbf{x}\rangle}{\parallel\varphi_{k-1}\parallel^2}
\end{split}
\end{equation}
the square of which becomes
\begin{align}
|\langle\gamma_i^{(k)}&, \varepsilon_{k-1}(\mathbf{x})\rangle|^2=\nonumber\\
&|\langle\gamma_i^{(k-1)},\varepsilon_{k-2}(\mathbf{x})\rangle|^2+\displaystyle\frac{|\langle\varphi_{k-1},\psi_i\rangle|^2\cdot|\langle\varphi_{k-1},\mathbf{x}\rangle|^2}{\parallel\varphi_{k-1}\parallel^4}\nonumber\\
&-2\langle\gamma_i^{(k-1)},\varepsilon_{k-2}(\mathbf{x})\rangle\cdot\frac{\langle\varphi_{k-1},\psi_i\rangle\cdot\langle\varphi_{k-1},\mathbf{x}\rangle}{\parallel\varphi_{k-1}\parallel^2}\label{sqr}
\end{align}
By applying $\eta_k(\mathbf{x})=\langle\varphi_{k-1},\mathbf{x}\rangle\varepsilon_{k-2}(\mathbf{x})$, Eq. \ref{sqr} can be re-formulated into
\begin{equation}
\begin{split}
&|\langle\gamma_i^{(k)},\varepsilon_{k-1}(\mathbf{x})\rangle|^2=\displaystyle|\langle\gamma_i^{(k-1)},\varepsilon_{k-2}(\mathbf{x})\rangle|^2\\
&-2\langle\psi_i,\eta_k(\mathbf{x})\rangle\cdot\frac{\langle\varphi_{k-1},\psi_i\rangle}{\parallel\varphi_{k-1}\parallel^2}+\frac{|\langle\varphi_{k-1},\psi_i\rangle|^2|\langle\varphi_{k-1},\mathbf{x}\rangle|^2}{\parallel\varphi_{k-1}\parallel^4}
\end{split}
\end{equation}
As above, it is derived that
\begin{equation}\label{opt}
\begin{split}
L&_k(\psi_i)=\frac{\parallel\gamma_{i}^{(k-1)}\parallel^2}{\parallel\gamma_i^{(k)}\parallel^2}L_{k-1}(\psi_i)\\
&-2\cdot\frac{\langle\psi_i,\varphi_{k-1}\rangle\cdot\langle\psi_i,\mathbf{I}_k\rangle}{\parallel\varphi_{k-1}\parallel^2\parallel\gamma_i^{(k)}\parallel^2}
+\frac{S_k|\langle\psi_i,\varphi_{k-1}\rangle|^2}{\parallel\varphi_{k-1}\parallel^4\parallel\gamma_i^{(k)}\parallel^2}
\end{split}
\end{equation}
where $\mathbf{I}_k=\frac{1}{N_f}\sum_{j=1}^{N_f}\eta_k(\mathbf{f}_j)-\frac{\lambda}{N_b}\sum_{j=1}^{N_b}\eta_k(\mathbf{b}_j)$
and $S_k=\frac{1}{N_f}\sum_{j=1}^{N_f}|\langle\varphi_{k-1},\mathbf{f}_j\rangle|^2-\frac{\lambda}{N_b}\sum_{j=1}^{N_b}|\langle\varphi_{k-1},\mathbf{b}_j\rangle|^2$.
By substituting the notations defined in Prop. 1, Eq. \ref{opt} becomes equivalent to
\begin{equation}\label{final}
\begin{split}
&L_k(\psi_i)=\\
&\frac{1}{d_i^{(k)}}\left[d_i^{(k-1)}L_{k-1}(\psi_i)-2\frac{\beta_i^{(k)}}{u_{k-1}}\langle\psi_i,\mathbf{I}_k\rangle+\left(\frac{\beta_i^{(k)}}{u_{k-1}}\right)^2S_k\right]
\end{split}
\end{equation}
\end{proof}

\end{document}